\documentclass[floatfix,aps,pra,twocolumn,showpacs,superscriptaddress,10pt,nofootinbib]{revtex4-2}

\usepackage{graphicx}
\usepackage{dcolumn}
\usepackage{bm}

\usepackage{subcaption}
\usepackage{float}
\usepackage{hyperref}
\usepackage{ragged2e}
\usepackage{amsmath}
\usepackage{xcolor}

\begin{document}

\title{MixFunn: A Neural Network for Differential Equations with Improved Generalization and Interpretability}

\author{Tiago de S. Farias}
\email{tiago.farias@ufscar.br}
\affiliation{Physics Department, Federal University of São Carlos, 13565-905, São Carlos, SP, Brazil}

\author{Gubio G. de Lima}
\affiliation{Physics Department, Federal University of São Carlos, 13565-905, São Carlos, SP, Brazil}

\author{Jonas Maziero}
\affiliation{Physics Department, 
Federal University of Santa Maria, 97105-900,
Santa Maria, RS, Brazil}

\author{Celso J. Villas-Boas}
\affiliation{Physics Department, Federal University of São Carlos, 13565-905, São Carlos, SP, Brazil}

\begin{abstract}
    We introduce MixFunn, a novel neural network architecture designed to solve differential equations with enhanced precision, interpretability, and generalization capability. The architecture comprises two key components: the mixed-function neuron, which integrates multiple parameterized nonlinear functions to improve representational flexibility, and the second-order neuron, which combines a linear transformation of its inputs with a quadratic term to capture cross-combinations of input variables. These features significantly enhance the expressive power of the network, enabling it to achieve comparable or superior results with drastically fewer parameters and a reduction of up to four orders of magnitude compared to conventional approaches. We applied MixFunn in a physics-informed setting to solve differential equations in classical mechanics, quantum mechanics, and fluid dynamics, demonstrating its effectiveness in achieving higher accuracy and improved generalization to regions outside the training domain relative to standard machine learning models. Furthermore, the architecture facilitates the extraction of interpretable analytical expressions, offering valuable insights into the underlying solutions.
\end{abstract}

\maketitle


\section{Introduction}

Differential equations lie at the heart of modeling a vast array of physical systems, spanning disciplines from classical mechanics, describing the motion of planets, rockets, and collisions, to quantum mechanics, where they govern the dynamics of molecules and elementary particles. These equations not only provide a framework for understanding fundamental natural phenomena but also serve as powerful tools for simulating complex physical processes. In engineering, differential equations are equally essential, used to model systems in fluid dynamics, structural analysis, electrical circuits, and thermodynamics. Through these applications, we can gain deeper insights into system behavior, enabling more accurate predictions and optimizations in scientific and engineering contexts. Consequently, differential equations are indispensable to scientific inquiry and technological advancements, forming a foundational basis for many of the technologies we rely on today \cite{ellahi_recent_2018, betounes_differential_2010}
.

A differential equation is a mathematical expression that involves the derivatives of a function with respect to its variables. Solving such equations entails finding the function that satisfies the equation for the given variables. The origins of differential equations trace back to the inception of calculus by Newton and Leibniz \cite{history}. Over the centuries, extensive research has been conducted on methods to solve these equations \cite{denis2020overview}.

Analytical techniques aim to derive exact or highly precise solutions in the form of mathematical functions. These solutions are particularly valuable as they allow for a deeper understanding of physical systems by enabling the analysis of the solution. However, these methods are limited in scope, as they can only be applied to relatively simple differential equations, which often do not represent the complexity of real-world physical phenomena \cite{xu2020, Ardourel2017}.

On the other hand, numerical methods offer a broader range of applicability by providing approximate solutions to more complex differential equations. These methods compute the values of the function at selected points within a domain of interest, making them suitable for various practical problems. Despite their versatility, numerical methods can be computationally expensive and often lack both the generalization capabilities and interpretability that are characteristic of analytical solutions. As a result, while they provide valuable numerical approximations, the underlying insights into the behavior of the system are often more challenging to extract \cite{review_diff_1, review_diff_3}.

Artificial neural networks (ANNs) have emerged as one of the most influential and versatile machine learning algorithms, demonstrating remarkable success in solving various problems across different domains. From image recognition \cite{alexnet} and natural language processing \cite{vaswani_attention_2017} to protein folding \cite{jumper_highly_2021}, neural networks have revolutionized these fields by providing highly accurate and scalable solutions \cite{dargan_survey_2020}. The key to their effectiveness lies in their expressivity: Their capacity to represent complex functions and relationships within data \cite{universal_1989}.

The expressivity of a neural network refers to the complexity it can model, which is closely tied to various factors such as the choice of activation functions, the depth and architecture of the network, and its arithmetic capabilities. More expressive networks, with the right design, can approximate highly complex functions, making them suitable for tackling traditionally difficult problems with analytical or numerical methods. However, this expressivity also introduces challenges, such as the need for extensive computational resources and the risk of overfitting when training on limited data. Thus, striking the right balance between network expressivity and generalization is a central concern in developing and applying neural networks across different fields.

The design choices made when constructing a neural network, including the selection of activation functions, network architecture, and other hyperparameters, have a profound impact on the ability of the model to solve a given problem effectively. These choices influence the capacity of the model to learn complex patterns, generalize to unseen data, and converge efficiently during training. While some of these decisions can be guided by theoretical analyses, such as understanding the mathematical properties of activation functions or the expressivity of deep networks, they can also be informed by empirical observations \cite{ithapu2017architecturalchoicesdeeplearning, philipp2021nonlinearitycoefficientpractical, 2021nika}.

In this work, we explore and combine two types of neurons for differential equations: the mixed-function and the second-order neurons. These neurons are designed to enhance the expressivity of neural networks in representing complex functions. These neurons form the foundation of two new models: the Mixed-Function Neural Network (MixFunn) and the Second-Order Mixed-Function Neural Network (Mix2Funn). By allowing each neuron to use multiple activation functions and incorporating relationships between inputs or other neurons, these models expand the arithmetic capabilities of standard networks.

A key advantage of our proposed models is their ability to represent complex functions with a drastically reduced number of parameters, up to four orders of magnitude fewer than conventional approaches. This substantial reduction not only diminishes computational costs but also facilitates the extraction of interpretable analytical expressions, thereby enhancing generalization and interpretability. The second-order neurons in Mix2Funn, in particular, introduce interactions between inputs, further bolstering the capacity of a model to capture higher-order dependencies. We demonstrate that these models outperform standard neural networks in solving differential equations, particularly in applications drawn from classical mechanics, quantum mechanics, and fluid dynamics.

The remainder of this article is organized as follows. In Section II, we review related papers that explore the application of neural networks to solving differential equations, highlighting their strengths, limitations, and key properties. Section III introduces our proposed methodology, detailing the design and functionality of the mixed-function and second-order neurons, as well as the regularization techniques that enhance the effectiveness of these models. In Section IV, we present and analyze the results of applying our proposed models to four distinct differential equation problems, discussing their performance and broader implications. Finally, in Section V, we provide concluding remarks, summarizing our contributions and outlining potential directions for future research.

\section{Related work}

Physics-Informed Neural Networks (PINNs) \cite{pinn_paper1, pinn_paper2, physics-informed_2019} integrate differentiable programming with differential equations to approximate the solution of complex physical systems. 
The core idea behind PINNs is to use the output of a neural network as a solution to a given differential equation and take advantage of the differentiable computational graph to enable differentiation. They approximate the solution to a differential equation by taking the input variables of the system, denoted as $\textbf{v}$, and feeding them into a neural network. The output of the network, $u(\textbf{v})$, represents the (approximate) solution to the differential equation. The training of PINNs revolves around minimizing one or more loss functions specifically designed to incorporate both the differential equation and any other constraint of the problem, ensuring the solution adheres to the underlying physics.

The first key component of the training process is the residual loss, which is constructed by substituting the output of the neural network into the differential equation and minimizing the resulting error. Let the differential equation be expressed as $\mathcal{D}\{u_{true}(\textbf{v})\} = 0$, where $u_{true}(\textbf{v})$ is the true solution and $\mathcal{D}$ is the differential operator. In PINNs, the neural network provides an approximation $u(\textbf{v})$, which is used instead of the exact solution. Given that an untrained neural network is likely to produce a solution with significant error, the residual becomes $\mathcal{D}\{u(\textbf{v})\} = \epsilon$, where $\epsilon$ represents the error or deviation from the true solution. To train the network, this residual must be minimized, often using the $L_2$ norm to quantify the error:
\begin{equation}
	\mathcal{L}_{residual} = ||\mathcal{D}\{u(\textbf{v})\}||_2^2.
\end{equation}

However, the residual loss alone is insufficient to approximate a differential equation's solution accurately. Without the inclusion of initial or boundary conditions, a differential equation has infinitely many possible solutions, including trivial solutions such as $u=0$, which do not necessarily reflect the true behavior of the system. This lack of specificity can impede the optimization process and lead the neural network to converge on undesirable or non-physical solutions.

A second component is introduced into the loss function to address this issue: the initial and/or boundary conditions ($IC/BC$) associated with the problem. These conditions provide essential constraints that the solution must satisfy at specific points in the domain. The $IC/BC$ loss is formulated by minimizing the discrepancy between the known conditions of the system and the predictions of the neural network at the corresponding points in the domain. This term ensures that the neural network adheres to the physical constraints imposed by the problem, thereby guiding the optimization towards a valid and meaningful solution. The loss for the initial and boundary conditions is typically given by:
\begin{equation}
	\mathcal{L}_{IC/BC} = \sum_k ||u(v_k) - u_{true}(v_k)||_2^2,
\end{equation}
where $v_k$ represents the points at which the initial or boundary conditions are specified, $u_{true}(v_k)$ is the known condition, and $u(v_k)$ is the output predicted by the network at those points. By incorporating this term, the optimization process is steered toward solutions that satisfy the differential equation and conform to the necessary initial and boundary conditions.

When additional information about the system is available in the form of known solutions at specific points, it can be incorporated into the training process through a data loss term. This loss, defined as the squared difference between the network's predictions and the known values at selected locations, helps guide the model toward the correct solution more efficiently. By constraining the network to fit these known data points, data loss accelerates convergence, reduces the risk of local minima, and improves overall accuracy. This approach is particularly useful in scenarios where partial solutions or experimental measurements are available, enhancing the reliability of the network.

Neural Differential Equations (NDEs) \cite{nde} approximate the derivatives of non-temporal variables instead of directly solving differential equations, embedding this approximation into a numerical solver. Unlike PINNs, which estimate solutions over the entire domain, NDEs learn the functional form of the derivative, allowing traditional integration methods to compute the solution over time. This approach is beneficial for modeling complex dynamics where explicit equations are unavailable. Specialized variants, such as Hamiltonian Neural Networks (HNNs) and Lagrangian Neural Networks (LNNs) \cite{hamiltonian_nn, cranmer2020lagrangian}, incorporate physical laws like energy and momentum conservation, ensuring physically consistent and interpretable models for dynamical systems. Complementary to these methods, the Sparse Identification of Nonlinear Dynamics (SINDy) model \cite{brunton2016discovering} employs sparse regression to discover governing equations directly from time-series data, revealing the underlying structure of complex dynamics by using a set of predefined functions.

In the context of Physics-Informed Neural Networks, artificial neural networks typically employ a single type of nonlinear function as the activation for each neuron. Common activation functions, such as the logistic sigmoid or hyperbolic tangent, are frequently used due to empirical success across various problem domains. Studies have shown that these functions perform well in practical applications, often yielding better results when solving complex tasks like image classification and pattern recognition \cite{krizhevsky_imagenet_2012, NEURIPS_relu}. However, the choice of a single, fixed activation function for all neurons can limit the interpretability of the network. Since the activation functions do not carry explicit information about the problem being solved, extracting meaningful insights about individual neurons or their contributions to the overall model becomes difficult.

Moreover, using a uniform activation function across the network can restrict the expressivity of the model, as it reduces the diversity of functional transformations that the network can represent. This limitation has led to growing interest in architectures that expand the variety of activation functions used within a network. The integration of multiple activation functions \cite{mixed_function, mixed2} addresses these limitations by combining multiple nonlinear activation functions within a single network. In this approach, each neuron or layer can be assigned a different activation function, allowing for a richer and more diverse representation of the data. By increasing the heterogeneity of the activations, a neural network with multiple activation functions enhances the expressivity of the model, enabling it to approximate more complex functions with fewer neurons and parameters. This flexibility also opens up the potential for improved generalization, as the network can adapt more effectively to the specific characteristics of the problem.

Furthermore, combining multiple activation functions contributes to better interpretability by providing a more granular understanding of the role each neuron plays in the network. Since different neurons can use distinct activation functions, it becomes possible to analyze their individual contributions in terms of their specific nonlinearities, offering a clearer picture of how the network processes and transforms the input data. This capability makes this type of model particularly promising for applications that require both high expressivity and interpretability, such as those involving physical systems.

SIREN (Sinusoidal Representation Networks) \cite{sirenpaper} is an architecture leveraging periodic activation functions for implicit neural representations. Unlike traditional multi-layer perceptrons (MLPs) with piecewise linear or smooth non-periodic activations, SIRENs are designed to model complex signals and their derivatives with high fidelity. It was demonstrated that the effectiveness of SIRENs in representing images, wavefields, and 3D shapes of objects, as well as solving boundary value problems such as the Poisson and Helmholtz equations \cite{sirenpaper}. SIRENs achieve fast convergence and precise modeling of spatial and temporal derivatives by incorporating a principled initialization scheme.

Parametric Rectified Linear Unit (PreLU) \cite{prelu_paper} is an activation function that adapts the slope of the negative part of the function through learnable parameters. This extension of the traditional ReLU activation enhances model flexibility while maintaining computational efficiency and mitigating overfitting risks.

The second generation of neuron models, which evolved from the original perceptron architecture introduced by Rosenblatt \cite{rosenblatt_perceptron:_1958}, extends the basic computational principles used in artificial neural networks. In the classical perceptron model, a neuron processes information by performing a linear operation on the input, a weighted sum of the inputs, followed by a nonlinear activation function. This simple yet powerful approach forms the basis of most neural network architectures today, including deep learning models \cite{lecun_deep_2015}. However, while this model has been highly effective in many applications, it has inherent limitations regarding its expressivity, as it only considers linear combinations of inputs.

Including product terms in the input data can enhance the expressive power of the machine learning algorithms \cite{gen_high1, gen_high2, gen_high3}. For instance, in polynomial regression, the ability to model quadratic terms allows for better approximation of curved data patterns \cite{harring_comparison_2012}, and they can be beneficial in applications where interactions between features are crucial, such as in physics simulations \cite{mcmahon_linear_2018, quadratic_bending}, financial modeling \cite{blanc2015quadratichawkesprocessesfinancial, Diaz2023}, and complex pattern recognition tasks \cite{liu_quadratic_2005, mckenzie_syntactic_1993}.

Neurons that capture higher-order correlations between inputs have a long history in the literature \cite{high_order_review}. Sigma-Pi neurons \cite{sigma_pi}, for instance, represent an advancement over the first-order perceptron by introducing a parameterized product term that accounts for interactions between different input variables or neurons. In contrast to the linear combination of inputs of the perceptron, higher-order neurons include multiplicative terms that capture the relationships among the inputs. This additional nonlinearity allows the network to model more complex interactions, enhancing the ability of the neuron to represent intricate patterns and dependencies in the data.

The self-attention mechanism in Transformers \cite{attention} is a powerful approach for capturing dependencies between input elements, making it central to modern neural network architectures. Unlike traditional sequence models, self-attention computes a weighted representation of the entire input sequence, enabling the model to focus on the most relevant parts for each element. It achieves this by transforming the input into queries, keys, and values, using attention scores, normalized via softmax, to determine the contribution of each element. This dynamic mechanism effectively captures local and global dependencies, enhancing the model's context-awareness.

While including these multiplicative terms does increase the number of parameters in the model, it can also improve the arithmetic capabilities of the neurons. By accounting for interactions between inputs, higher-order neurons offer a richer and more flexible framework for function approximation. This increased expressivity enables the neural network to model more complex systems with fewer neurons, as each neuron can encode a broader range of functional transformations than its first-order counterpart.

Moreover, this type of neuron is advantageous in scenarios where interactions between variables are critical, such as in modeling physical systems, higher-dimensional data spaces, or other domains where relationships between inputs are nonlinear and complex. Although this parameter increase can lead to higher computational costs, the benefits of model performance and accuracy can outweigh the added complexity. Thus, higher-order neurons offer a powerful tool for enhancing neural networks' expressivity and representational capacity, making them well-suited for solving more challenging problems.

Kolmogorov-Arnold Networks (KANs) \cite{liu2024kan} represent another approach to neural network design, inspired by the Kolmogorov-Arnold representation theorem. Unlike traditional MLPs, which use fixed activation functions at nodes, KANs employ learnable activation functions along edges, parameterized as splines. This design choice eliminates linear weight matrices and enables KANs to optimize univariate functions while leveraging compositional structures adaptively. By combining the strengths of MLPs and splines, KANs address challenges such as the curse of dimensionality, achieving better scaling laws and accuracy in function fitting tasks, particularly for data with compositional properties.

Eigenvalue problems are fundamental to many areas of physics and mathematics, particularly in quantum mechanics, where they arise in studying wavefunctions and energy levels. These problems typically involve differential equations with boundary conditions that must be satisfied by the solution. Traditional numerical methods for solving eigenvalue problems include the shooting method, which transforms the boundary value problem into an initial value problem. The shooting method \cite{scheiber2022adjointshootingmethodsolve, singh2022shootingmethodsolvingtwopoint} iteratively adjusts the initial conditions to match the desired boundary conditions at the other end of the domain, converging to the correct eigenvalues and eigenfunctions. While effective, this method can be computationally intensive and sensitive to initial guesses, particularly for high-dimensional or complex systems.

Ref. \cite{jin2022physicsinformed} proposes an approach for solving quantum eigenvalue problems using Physics-Informed Neural Networks (PINNs). Their method introduces two physics-based loss functions: ortho-loss, which encourages the discovery of pairwise orthogonal eigenfunctions, and norm-loss, which ensures normalization of eigenfunctions to avoid trivial solutions. These enhancements allow the network to learn eigenvalues and eigenfunctions simultaneously directly, without the need for data, relying solely on the network's predictions. Additionally, the architecture incorporates embedded symmetries to accelerate convergence for systems with even or odd eigenfunctions, further improving efficiency and accuracy. This framework was demonstrated on standard quantum mechanics problems, including the finite square well, multiple finite square wells, and the Hydrogen atom.

\section{Methods}

In this section, we describe the core methodologies underlying the MixFunn architecture, focusing on its components and regularization techniques. MixFunn combines the expressive power of second-order neurons with the flexibility of mixed-function neurons to address the challenges of solving differential equations. To further enhance the performance of the model and prevent overfitting, we incorporate regularization techniques such as dropout \cite{dropout_paper}, pruning \cite{frankle2018the}, and softmax-based normalization \cite{softmax_paper}, promoting simpler and more interpretable solutions.

\subsection{Second-order neuron}

In the context of neural networks, a neuron can be considered as a parameterized function of its inputs. This function, denoted as $s(\textbf{x};\theta)$, where $\textbf{x}$ are its inputs and $\theta$ the parameters, can take various forms subject to certain restrictions. Since the exact form of this function is unknown, we can approximate it through a series expansion. The simplest approximation is the zero-order neuron, represented as $s(\textbf{x};\theta)=b,$ where $b$ is a scalar bias parameter. This zero-order neuron does not depend on its inputs and has limited learning capacity.

The next level of expansion is the first-order neuron, expressed as the linear mapping $s=\textbf{W}\textbf{x}+b$, where $\textbf{W}$ is the weight matrix. Combined with a subsequent nonlinear activation function, this linear model forms the foundation of most deep learning models. First-order neurons have been highly successful in solving a wide range of tasks due to their simplicity and efficiency in capturing linear relationships within the data.

However, the linear nature of first-order neurons presents several limitations. Firstly, they can only capture linear relationships between inputs, often insufficient for modeling complex data. This limitation is typically addressed by stacking multiple neurons with nonlinear activation functions in deeper architectures. Secondly, in the case of multidimensional data, the linear mapping only captures individual data features but fails to capture the interactions between these features. 
Again, this can be mitigated by using multiple nonlinear neurons. For example, consider the task of learning to multiply two numbers $x_1$ and $x_2$. A neural network composed solely of first-order neurons struggles to generalize this multiplication beyond the training domain, as it cannot inherently model the multiplicative interaction between $x_1$ and $x_2$.

\begin{figure}[t]
\includegraphics[width=0.3\linewidth]{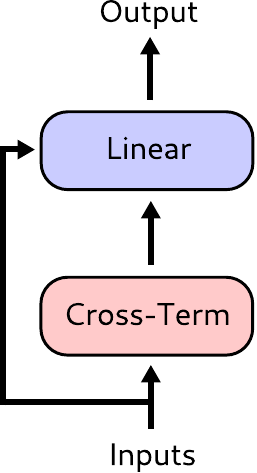}
\caption{\justifying Schematic representation of a second-order neuron in a neural network. 
The input elements are cross-correlated up to second-order, then the cross-terms and the inputs are summed and acted on by a linear transformation. 
}
\label{fig:secondorderneuron}
\end{figure}

We can extend the approximation to second-order neurons to capture more complex patterns and relationships in the data, as shown in Figure \ref{fig:secondorderneuron}. Second-order neurons account not only for the weighted inputs but also the weighted interactions between pairs of inputs. The second-order neuron is defined as:
\begin{equation}
	s = b + \sum_i^N w_i x_i + \sum_{i}^N \sum_{j}^N u_{ij} x_i x_j.
\end{equation}

In this equation, $b$ is the bias term, $w_i$ are the weights for the individual inputs $x_i$, and $u_{ij}$ are the weights representing the interactions between pairs of inputs $x_i$ and $x_j$, summed over their $N$ inputs. This formulation allows the neuron to capture quadratic relationships within the input data, thereby enabling the modeling of more complex interactions beyond the capabilities of first-order neurons. For instance, whereas first-order neural networks can only approximate the multiplication of two numbers, second-order neural networks can directly learn this operation with a comparatively small set of parameters.

In vector form, the output of a second-order neuron can be expressed as:

\begin{equation}\label{equ:second_order}
    s = b + \textbf{W}\textbf{x} + \textbf{U} \textbf{x} \textbf{x}^T,
\end{equation}
where $b$ is the bias parameter, \textbf{W} is the weight matrix for the first-order terms, with size $1\times N$, and \textbf{U} is the weight tensor for the second-order terms, with size $1 \times N \times N$. The term \textbf{x}\textbf{x}$^T$, with size $N \times N$, forms a matrix where each element $(i,j)$ represents the product $x_i x_j$, effectively capturing all pairwise interactions between the input features. The tensor \textbf{U} is then contracted with this matrix, meaning that the corresponding elements of \textbf{U} and \textbf{x}\textbf{x}$^T$ are multiplied element-wise and summed, to yield a scalar that quantifies the contribution of the second-order interactions.

It is important to recognize that the outer product \textbf{x}\textbf{x}$^T$ is symmetric, since $x_i x_j = x_j x_i$, which introduces redundant computations and can lead to inefficiencies. To mitigate this issue, we restrict the computation to one triangular portion (upper or lower) of the matrix. Accordingly, we can reformulate the second-order neuron as follows:
\begin{equation}
    s = b + \textbf{W}\textbf{x} + \textbf{U} vec(tril(\textbf{x} \textbf{x}^T)),
\end{equation}
where $tril()$ extracts the lower triangular part of the matrix and $vec()$ vectorizes it into a one-dimensional array. As a result, the weight tensor \textbf{U} is redefined as a matrix of dimensions $1\times N(N-1)/2$. This adjustment reduces the number of elements from $N^2$ to $N(N-1)/2$, eliminating redundancy and enhancing computational efficiency. Although the memory complexity remains $O(N^2)$, the nearly $50\%$ reduction in stored and processed elements leads to computational savings.

While second-order neurons introduce nonlinearity through their quadratic terms, this quadratic form alone does not provide the full expressive power required for complex function approximation. Although quadratic functions can capture more intricate relationships than linear functions, they are still limited in their representational capacity. Specifically, the quadratic form, which resembles polynomial functions, does not satisfy the conditions of the universal approximation theorem \cite{LESHNO1993861}. Therefore, following them with an additional nonlinear activation function is essential to harness the full potential of second-order neurons and achieve universal approximation.

\subsection{Mixed-function neuron}

\begin{figure}[t]
\includegraphics[width=0.3\linewidth]{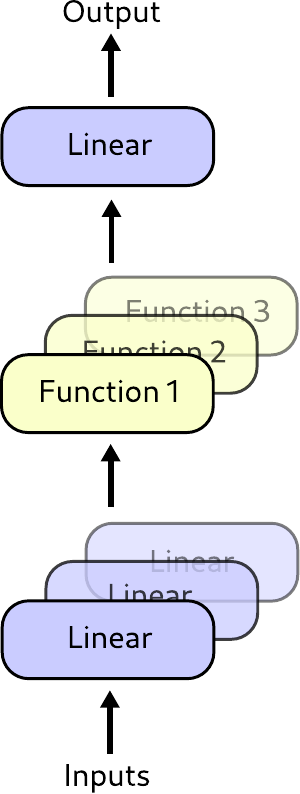}
\caption{\justifying Diagram of a mixed-function neuron in a neural network. This neuron utilizes a set of diverse nonlinear functions $\{f_i(s_i)\}^Q$, where $Q$ represents the number of functions (three in this figure). The output is a linear combination of these activation functions, given by $a=\sum_i^Q w_i f_i(s)$, with $w_i$ as the weights for each function $f_i(s_i)$ and $s_i$ as the input, which can be the data inputs, first-order (as shown in this figure) or second-order neurons. The selection of the activation functions is based on their differentiability, finiteness within the domain of interest, and the finiteness of their derivatives up to the highest order present in the differential equation being solved.}
\label{fig:mixedfunction}
\end{figure}

Traditional neurons typically use a single activation function, such as the rectified linear unit \cite{2019relu} or hyperbolic tangent. A mixed-function neuron uses a group of diverse nonlinear activation functions, denoted by the set $\{f\}^Q$, where $Q$ represents the number of these functions. The output of a mixed-function neuron is defined as a linear combination of these nonlinear functions. Mathematically, the output $a$ of such a neuron is expressed as:
\begin{equation}
	a = \sum_{i=1}^Q w_i f_i (s_i),
\end{equation} 
where $w_i$ are the weights associated with each function $f_i(s_i)$, and $s_i$ is the input to the neuron. The selection of the functions $f_i(s_i)$ is determined based on three critical requirements. Firstly, the chosen functions must be differentiable, ensuring the application of gradient-based optimization methods. Secondly, the functions must be finite within the domain of interest to maintain numerical stability and meaningful outputs. Lastly, the derivatives of these functions, up to the highest order present in the differential equation under consideration, must also be finite across the same domain. Figure \ref{fig:mixedfunction} shows a graphical representation of the mixed-function neuron.

In this work, we propose using mixed-function neurons to approximate solutions of differential equations within the domain of physics. A key concept in the effectiveness of neural networks is inductive bias, which refers to the assumptions or prior knowledge incorporated into the model to solve specific problems more efficiently. Inductive bias influences the design choices in neural network architecture and the selection of functions, ultimately guiding the learning process towards more plausible solutions based on the inherent characteristics of the problem. For example, convolutional neural networks \cite{conv1, conv2, conv3} exhibit an inductive bias that is particularly well-suited for computer vision tasks. This is because the convolution operator effectively captures local spatial hierarchies and patterns within image data, such as edges and textures, enhancing the network's ability to recognize and interpret visual information.

Similarly, in the context of mixed-function neurons, selecting appropriate nonlinear functions is crucial for accurately capturing the dynamics of physical systems. Functions that frequently appear in physical phenomena are particularly well-suited for this purpose. Oscillatory and decay behaviors are common in many physical contexts, making trigonometric functions such as sine and cosine, as well as exponential functions, excellent candidates. The sine and cosine functions are particularly effective in modelling periodic phenomena, such as wave propagation and harmonic motion. In contrast, exponential functions are ideal for representing growth and decay processes, including radioactive decay and population dynamics.

Certain functions such as the inverse function $f(x)=1/x$ and the logarithm function $f(x)=\log(x)$ are also frequently encountered in the domain of physics. However, these functions often exhibit numerical instability or are undefined over portions of their domains, posing significant challenges in computational applications. For instance, the logarithm function is undefined for $x\le0$, leading to complications in calculations involving negative or zero values.

To mitigate these issues, one can adopt two primary strategies: restricting the domain or modifying the function to achieve a smoother approximation. Restricting the domain involves limiting the function's input to a well-behaved range. Alternatively, modifying the function can provide a more practical solution. For example, the logarithm function can be approximated by a well-defined and stable function across a broader domain. One such transformation is $f(x)=\log(k+ReLU(x))$, where $ReLU(x)$ denotes the Rectified Linear Unit function, defined as $ReLU(x)=\max(0,x)$, and $k$ is a small positive constant value. This modification ensures that the function is defined for all real numbers $x$, thereby enhancing numerical stability and maintaining the desired properties of the original function in the domain of interest.

When combined with second-order neurons, we extend this framework to create mixed-function second-order neural networks, which we refer to as \textit{Mix2Funn}. This hybrid approach leverages the expressive power of second-order interactions along with the flexibility of mixed-function neurons, enhancing the model's ability to capture complex patterns and interactions in the data. This architecture is particularly advantageous in domains where nonlinear relationships and interactions between features are critical, such as in the solution of differential equations.

\subsection{Normalization of mixed-functions by softmax}

The linear combination of nonlinear functions in a mixed-function neuron involves independent weights that are optimized during training. Consequently, a mixed-function neuron can produce an output that is a combination of multiple functions simultaneously. However, in certain scenarios, it may be desirable for the neuron to effectively select a single function from the set rather than combining them. This requirement relates to Kolmogorov complexity \cite{complexity1, complexity2, complexity3}, a concept in information theory that quantifies the complexity of an object based on the length of the shortest possible description of that object in a fixed universal language. Kolmogorov complexity essentially quantifies the amount of information required to describe an object, the simpler the object, the lower its complexity.

Kolmogorov complexity is closely aligned with the philosophical principle of Occam’s razor \cite{occam}, which posits that the one with the fewest assumptions should be selected among competing hypotheses. Simpler solutions are generally preferred over more complex ones, as they are more likely to generalize well to new data. Applying Occam’s razor in the context of neural networks, we seek accurate and simpler models, avoiding unnecessary complexity that could lead to overfitting.

In the context of differential equations, this implies that shorter, more straightforward solutions are more likely to be accurate than longer, more complex ones. We can design the mixed-function neuron to favor selecting a single function from its set during training to enforce this principle. One method to achieve this is by constraining the weights of the linear combination.

To enforce constraints ensuring that weights  $w_i$ are non-negative and sum to one, the softmax function \cite{softmax_paper} is commonly employed. This function normalizes a vector of unnormalized parameters $\alpha_i$ into a probability distribution, facilitating the model's focus on specific functions within a set. The softmax function is defined as:
\begin{equation}
	w_i = \frac{e^{\alpha_i/T}}{\sum_j e^{\alpha_j/T}},
\end{equation}
where $T$ is a temperature hyperparameter that controls the smoothness of the output distribution, adjusting T influences the entropy of the distribution: higher values lead to a more uniform distribution, while lower values result in a sharper distribution where one function may dominate. This mechanism aligns with the concept of Kolmogorov complexity by promoting simpler solutions, potentially leading to more generalizable models for approximating differential equations.

An annealing strategy can be applied to the temperature parameter \cite{zhang2018heatedupsoftmaxembedding} to enhance training effectiveness. Starting with a high temperature allows the model to explore a broad range of functions due to a uniform weight distribution. Gradually decreasing the temperature encourages the model to converge towards selecting the most appropriate function, balancing exploration and exploitation during training. 

\subsection{Dropout}

Dropout \cite{dropout_paper} is a powerful regularization technique that prevents overfitting in neural networks by enabling the network to effectively search for and improve sub-networks within its architecture. This technique operates by randomly setting the activation of a subset of neurons to zero with a specified probability during the training phase. Consequently, during each training iteration, different sets of neurons are ``dropped out'' and do not contribute to the output of the network, nor do they participate in the gradient computation during backpropagation.

Mathematically, if the activation of a neuron is represented by $a_i$, dropout applies a mask $m_i$ sampled from a Bernoulli distribution with probability $p$ (the dropout rate), resulting in the modified activation $\tilde{a}_i = m_i a_i$, where $m_i \in \{0,1\}$ and $Pr(m_i=1)=1-p$, indicating the probability that a activation of a neuron is retained.

As training progresses, neurons are randomly excluded from the forward pass, forming numerous smaller sub-networks. This stochastic removal of neurons compels the network to learn redundant representations and distribute weights more broadly. The gradient updates during backpropagation are applied only to the active neurons, ensuring that the learning process continues effectively even in the presence of dropout. By blocking specific paths from the output to the input during the backward pass, dropout encourages the network not to rely on specific neurons and instead develop a more distributed and generalizable internal representation, reducing the risk of overfitting.

Mixed-function neural networks, whether they incorporate second-order neurons or not, can benefit from the application of dropout. This regularization technique facilitates the network's exploration of diverse combinations of functions, enhancing its ability to approximate solutions to differential equations. By randomly deactivating neurons during training, dropout prevents the network from becoming overly dependent on specific subsets of functions. This stochastic regularization potentially promotes the discovery of simpler, more robust solutions that generalize well to unseen data. Additionally, dropout enhances the model's flexibility by enabling it to explore different combinations of functions dynamically. This dynamic exploration is particularly valuable in solving differential equations, where the solution space can be highly complex and multi-faceted. Consequently, dropout ensures that the network remains versatile and capable of adapting to various functional forms.

This approach aligns with the principles discussed in the previous section regarding Kolmogorov complexity and Occam’s razor. By enforcing a regularization mechanism, dropout aids in pursuing simpler models. It encourages the network to avoid overfitting to the training data and instead find solutions that capture the underlying structure of the problem in a more generalized manner.

\subsection{Pruning}

Pruning is an effective regularization technique that can enhance the generalization capability of neural networks by reducing their complexity \cite{frankle2018the}. The essence of pruning lies in permanently removing parameters from the neural network based on predefined criteria. Various strategies for pruning exist, each tailored to achieve specific objectives. One commonly used approach is random pruning, which stochastically selects parameters to be removed based on a predefined probability. Alternatively, magnitude pruning targets parameters with values below a certain threshold, under the assumption that these parameters contribute less significantly to the performance of the model.

In practice, pruning techniques are typically implemented by applying a binary mask consisting of zeros and ones. This mask effectively nullifies the parameters corresponding to zero entries, creating a virtual pruning effect. After pruning is used in the optimization process, there are two primary strategies to improve the network:  reconstructing the neural network by removing the parameters associated with zeros in the mask, or leveraging sparsity techniques to execute the pruned network more efficiently without altering its structure.

In the context of mixed-function neural networks, where neurons combine multiple nonlinear activation functions parametrically, pruning plays a role in steering the network towards simpler solutions. However, applying pruning in this setting requires careful consideration of the unique contributions of different nonlinear functions. For example, some activation functions, such as the sine function, can influence the output of the neural network even at lower parameter values (e.g., $\cos{(0)}=1$). This characteristic can pose challenges for pruning techniques that rely on the magnitude of parameter contributions as a criterion for removal. To address this, pruning can be done at a more global level. Specifically, since mixed-function neurons involve parameterized linear combinations of distinct nonlinear functions, pruning these parameters effectively removes entire functions from the neuron, rather than targeting individual contributions within the linear combination. This global pruning approach ensures a more coherent reduction of complexity while preserving the functional integrity of the network.

\section{Results}

In this section, we apply our proposed technique to approximate solutions for four distinct problems: the damped and forced harmonic oscillator, the Burgers' equation, and the infinite quantum well. These problems were chosen for their diversity across different domains of application. The harmonic oscillator represents a fundamental model in classical mechanics and physics, often used to describe systems ranging from simple pendulums to molecular vibrations. The Burgers' equation, a canonical example in fluid dynamics, is a prototype for modeling shock waves and turbulence, making it highly relevant for engineering applications. Lastly, the infinite quantum well, a classic problem in quantum mechanics, poses an eigenvalue problem that is crucial for understanding quantum confinement and energy quantization. 

In addition to evaluating the performance of our technique within the training domain, we investigate the generalization capability of the models by comparing their approximated solutions to known analytical or numerical solutions of the systems. By assessing the generalization, we test the performance of the model on data points outside the training domain, examining its ability to predict solutions in regions it has not previously encountered accurately. This approach allows us to evaluate how well the model extrapolates beyond the initial conditions and constraints provided during training.

For each problem, we present a detailed specification of the corresponding differential equation, including the initial and boundary conditions, the defined training domain, and the domain used for testing generalization. To benchmark the performance of our approach, we also compare the results with the standard formulation of PINNs, which employ the hyperbolic tangent as the activation function with traditional first-order neurons. In all cases, we employ the Adam optimizer \cite{kingma2017adammethodstochasticoptimization}, with hyperparameters determined empirically through a systematic search. Specifically, a learning rate of $\eta=0.1$ is used for models employing our proposed MixFunn architecture, while $\eta=0.01$ is applied for the standard PINN models. Moreover, we found that exponential decay rates for the moment estimates, with $\beta_1=0.9$ and $\beta_2=0.9$, consistently yielded the best performance across all models.

For the MixFunn models, we selected seven distinct functions based on the inductive bias discussed earlier. These functions were chosen to enhance the ability of a network to approximate differential equation solutions by leveraging commonly observed functional forms in physical systems. The selected functions include sine, cosine, exponential of absolute value, exponential of negative absolute value, square root, natural logarithm, and the identity function.

This selection ensures that the network can represent various behaviors, including oscillatory dynamics, exponential growth and decay, and power-law relationships, providing greater flexibility in capturing complex patterns while maintaining interpretability. By incorporating these diverse functions, MixFunn introduces a structured approach to function selection, reducing the reliance on arbitrary activation choices and improving generalization to various differential equations.

All equations and parameter values are presented in a nondimensionalized form, which ensures consistency without the explicit use of physical units. This formulation allows users to adopt any unit system of their choice, provided that consistency is maintained across all parameters.

\subsection{Harmonic oscillator}

The first problem we investigate is the forced harmonic oscillator with damping \cite{oscillator}. This model is fundamental in capturing a wide range of oscillatory behaviors and serves as the basis for understanding many physical phenomena. It exhibits interesting phenomena such as resonance, damping, and criticality, among others. The differential equation governing a general harmonic oscillator is defined as:
\begin{equation}
	m\frac{d^2 x}{dt^2} + \gamma\frac{dx}{dt} + kx = F(t),
\end{equation}
where $x$ represents the position, $m$ the mass of the oscillator, $\gamma$ is the damping coefficient, $k$ is the elastic (spring) constant, and $F(t)$ is the external force, which can be a function of time. The term $m\frac{d^2 x}{dt^2}$ represents the inertial force, $\gamma\frac{dx}{dt}$ accounts for the damping force, and $ kx$ is the restoring force due to the spring. The external force $F(t) $ can vary with time, introducing complexities such as resonance when the driving frequency matches the natural frequency of the system. The forced harmonic oscillator with damping is a prototypical system in classical mechanics, providing insights into more complex dynamical systems. It illustrates key concepts such as natural frequency, damping ratio, and forced response, which are fundamental in studying vibrations, electrical circuits, and other oscillatory systems.

\subsubsection{Damped harmonic oscillator}

One of the fundamental cases in the study of harmonic oscillation is the damped harmonic oscillator, characterized by the absence of an external driving force, i.e., $F(t)=0$. Despite its simplicity, this system exhibits rich dynamical behavior, such as decaying oscillations, which makes it a suitable candidate for initial investigations using PINNs. In this work, we examine the damped harmonic oscillator under the specific conditions where the mass $m=1$, the spring constant $k=1$, and the damping coefficient $\gamma=0.1$. The initial conditions are set as $x(0)=1$ and $\dot{x}(0)=0$, representing an initial displacement with no initial velocity.

\begin{figure*}
    \centering
    \begin{subfigure}{0.33\textwidth}
    \centering
    \includegraphics[width=1\linewidth]{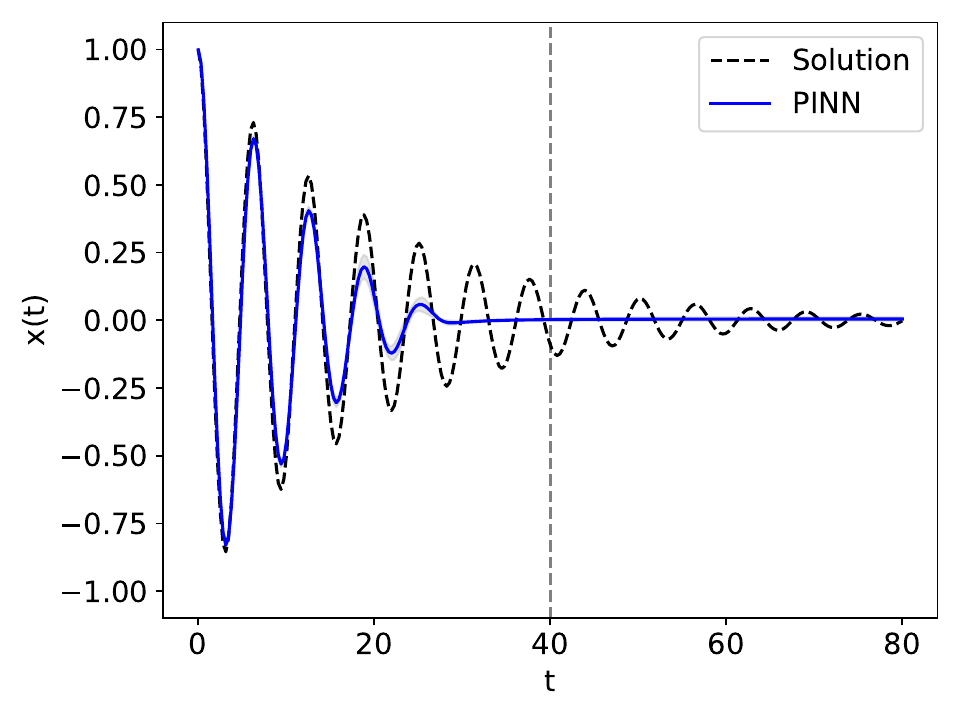}
        \caption{}
        \label{}
    \end{subfigure}%
        \begin{subfigure}{0.33\textwidth}
    \centering
    \includegraphics[width=1\linewidth]{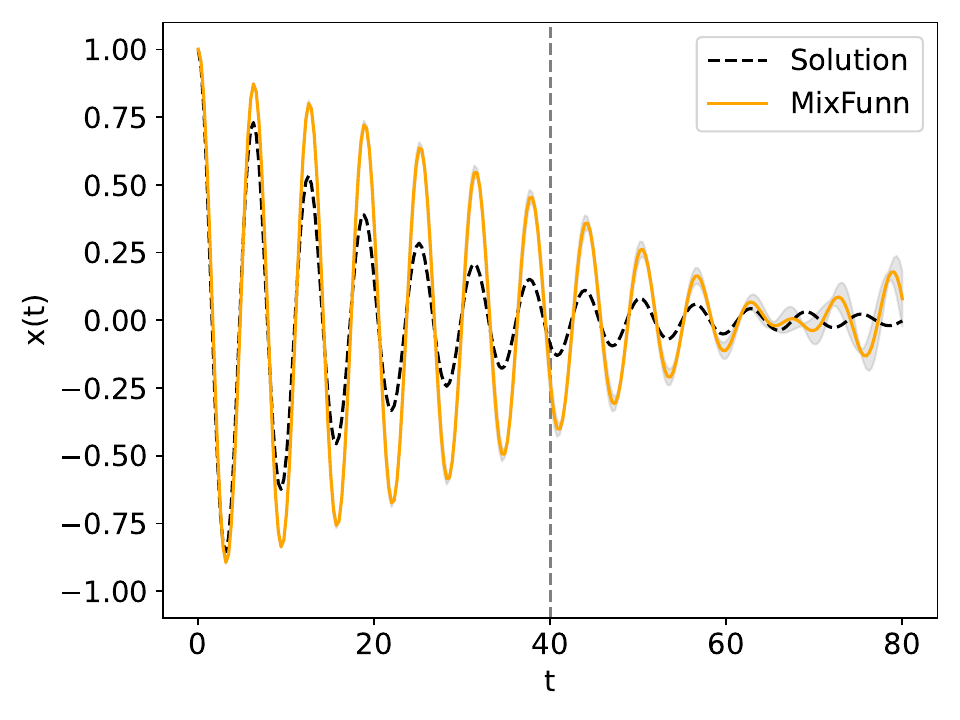}
        \caption{}
        \label{}
    \end{subfigure}
    \begin{subfigure}{0.33\textwidth}
    \centering
    \includegraphics[width=1\linewidth]{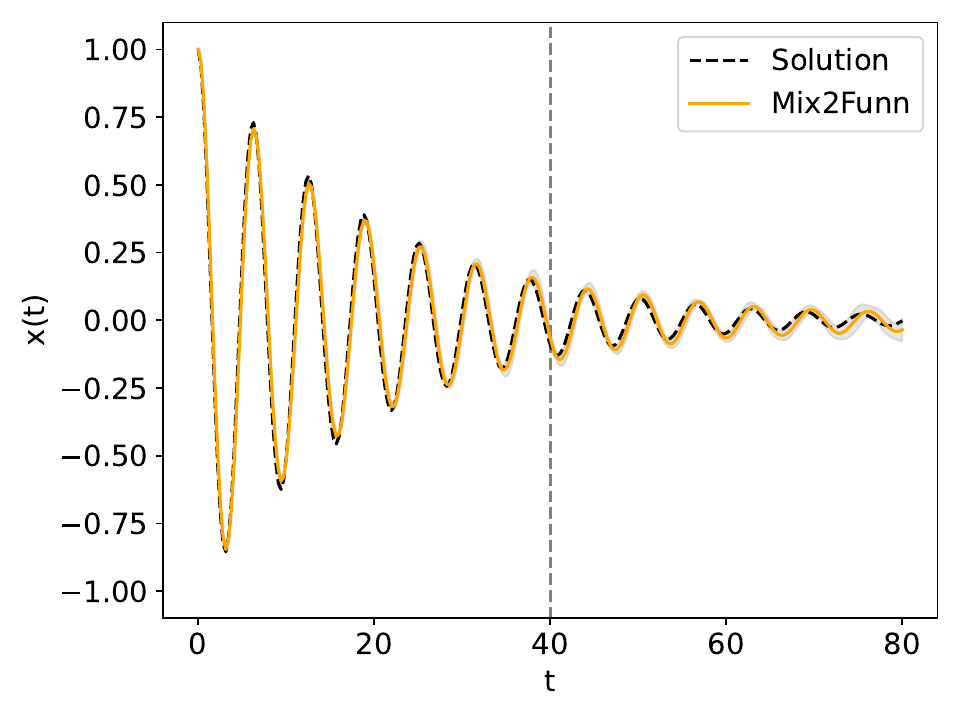}
        \caption{}
        \label{}
    \end{subfigure}
    \caption{\justifying Comparison of the approximated solutions for the damped harmonic oscillator using the standard (a) PINN, (b) MixFunn, and (c) Mix2Funn models. The standard PINN fails to capture the dynamics accurately within the training domain and struggles with generalization in the test domain. In contrast, the Mix2Funn models perform significantly better, successfully approximating the oscillatory behavior even for large values of $t$. The shaded regions around the predicted solutions represent the variance due to five different model initializations.}
    \label{fig:damped}
\end{figure*}

Figure  \ref{fig:damped} presents the results of the approximated solutions for the damped harmonic oscillator, comparing the performance of three models: the standard PINN, MixFunn, and Mix2Funn. The standard PINN shows an apparent failure to accurately approximate the solution, even within the training domain, whereas Mix2Funn performed significantly better in capturing the dynamics of the system.

When implemented without second-order neurons, the MixFunn model exhibits difficulty accurately capturing the solution values despite displaying decaying oscillatory behavior within the training domain. This limitation arises from the linear combination of nonlinear functions, which proves insufficient for identifying a parameter set that adequately represents the correct solution.

A potential explanation for the poor performance of the standard PINN lies in its difficulty in handling large input values. In deep learning, it is a common practice to normalize inputs to prevent saturation of the activation functions and to maintain adequate gradient flow. However, in the case of differential equations like the one governing the damped harmonic oscillator, normalization may not be feasible due to the inherent physical constraints of the problem. As a result, the standard neural network experiences saturation of its activation functions for large input values, leading to vanishing gradients that hinder effective optimization. This saturation significantly reduces the ability of the model to learn and approximate the solution accurately.

In contrast, the Mix2Funn model mitigates this issue by combining selected mixed functions as activation units. These mixed functions allow each network component to contribute meaningfully, even with large input values. As a result, the model maintains sufficient gradient magnitude and avoids the common pitfalls of standard activation functions. 

The standard PINN demonstrates a significant failure in generalizing the results in the test domain, producing a constant output for large values of $t$. This outcome can once again be attributed to the oversaturation of the neurons, which limits its ability to capture the dynamic behavior of the system as time progresses. In contrast, our proposed method, both the MixFunn and Mix2Funn, models exhibit better performance, accurately capturing the ongoing oscillatory dynamics of the damped harmonic oscillator even for large values of $t$.

Table \ref{table:damped} provides a detailed comparison of the mean training error, mean test error, and their respective standard deviations across the five different parameter initializations. Additionally, it highlights the lowest test error achieved across all initializations. This comparison shows that the Mix2Funn models achieve lower mean errors and consistently attain a much lower minimum test error than the standard PINN. 

\begin{table}
\fontsize{8pt}{8pt}
\begin{center}
\begin{tabular}{ | m{2.cm} | m{2.cm}| m{2.cm} | m{2.cm} |} 
  \hline
   Model & Training error & Test error & Lowest error \\ 
  \hline
  \hline
  PINN & \vspace{2pt} $1.43 \times 10^{-2} \pm 2.08 \times 10^{-3}$ &  $2.43 \times 10^{-3} \pm 7.74 \times 10^{-5}$ & $2.35 \times 10^{-3}$ \\
  \hline
    MixFunn & \vspace{2pt} $4.29 \times 10^{-2} \pm 1.90 \times 10^{-3}$ & $1.47 \times 10^{-2} \pm 4.94 \times 10^{-4}$ & $1.42 \times 10^{-2}$ \\ 
  \hline
  Mix2Funn &  \boldmath{\vspace{2pt} $2.68 \times 10^{-3} \pm 2.57 \times 10^{-3}$} & \boldmath{$1.04 \times 10^{-3} \pm 8.86 \times 10^{-4} $} & \boldmath{ $1.37 \times 10^{-4}$} \\ 
  \hline
\end{tabular}
\end{center}
\caption{\justifying Comparison of training and test errors, along with the lowest test error achieved, for the standard PINN, MixFunn, and Mix2Funn models on the damped harmonic oscillator problem. The results highlight better performance and generalization capability of the Mix2Funn model, with significantly lower errors than the standard PINN. Standard deviations indicate the variability due to different model initializations, with Mix2Funn achieving the lowest overall error across initializations.}
\label{table:damped}
\end{table}

\subsubsection{The role of data size}

In the previous cases involving the damped harmonic oscillator, we constrained the training domain by setting a fixed maximum value, allowing the test domain to extend beyond this range. This approach enables us to assess not only the ability of the model to approximate the solution within the training domain but also its capacity to generalize to unseen data outside the training domain. This balance between fitting the training data and generalizing to new inputs aligns with the broader concept of out-of-distribution generalization in machine learning models \cite{out_of_distribution1, out_of_distribution2}.

Several studies in the literature \cite{generalization1, generalization2, generalization3, generalization4, generalization5} suggest that the number of training data points plays a critical role in the generalization of a model, typically measured by its ability to predict unseen data accurately. Intuitively, a larger dataset provides more information for the model to learn from, thereby improving its performance on new or out-of-distribution examples. In physics-informed neural networks, the unseen data corresponds to the points outside the predefined training domain. This raises an important question: could increasing the size of the training domain improve the generalization capability of a model? By expanding the domain, the model might gain more exposure to the underlying physical dynamics, potentially enhancing its ability to generalize beyond the training region and solve the differential equation in regions where no explicit training data was provided. 

To evaluate the generalization capacity of the model across different training data sizes, we designed the experiment with a test range for all models, ensuring a fair comparison. Specifically, the test domain was fixed at  $t \in [80, 150]$ across all trials, regardless of the training domain size. For the training domains, we varied $T_{max}$, using $t$ values from the interval $[0,T_{max}]$, where $T_{max} \in \{10,20,40,60,80\}$. Given that the results from the previous section demonstrated the failure of the standard PINN to generalize effectively, we focused this evaluation exclusively on the Mix2Funn model. By testing across multiple values of $T_{max}$, we aim to identify the optimal balance between training domain size and model complexity for achieving the best generalization results.

\begin{figure}[t]
\includegraphics[width=1\linewidth]{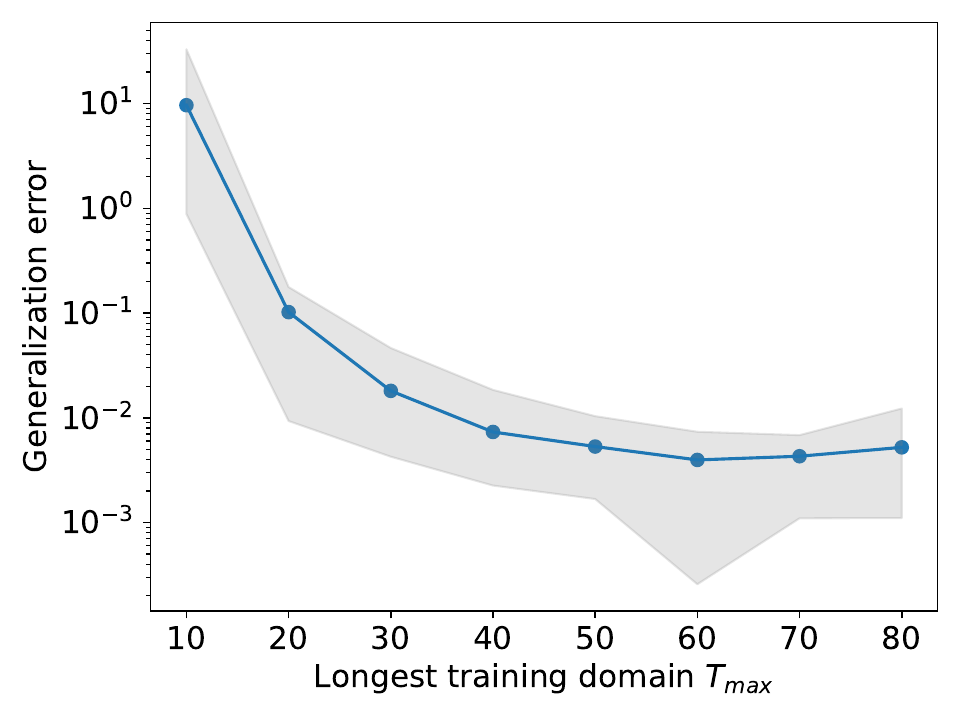}
\caption{\justifying Relationship between generalization error and the maximum value of the training domain $T_{max}$. The generalization error decreases as $T_{max}$ increases, indicating improved generalization capability. The shaded region indicates the variation in the average error due to different parameter initializations of the network.}
\label{fig:datasize_loss}
\end{figure}

Figure \ref{fig:datasize_loss} illustrates the relationship between the generalization error and the maximum value of the training domain, $T_{max}$. The results indicate that the generalization error decreases as $T_{max}$ increases. This suggests that providing the model with more information about the system, through a larger training domain, enhances its ability to generalize to unseen data outside the training range. This finding aligns with the broader understanding that a more comprehensive dataset enables the model to better capture the underlying dynamics of the differential equation, thus improving its predictive performance.

One possible explanation for this behavior is that the optimal solution of the neural network, that is, the solution that closely approximates the true solution of the differential equation, can be thought of as existing within a sub-network of the overall architecture. Expanding the training domain exposes the model to a broader variety of data, forcing the network to explore a larger portion of the solution space. In doing so, the network effectively ``searches'' for the correct sub-network by minimizing the influence of parameters associated with incorrect solutions. This process is akin to progressively reducing the contribution of sub-optimal parameters to zero, allowing the model to converge towards the correct solution.

\subsubsection{Extracting analytical solutions}

Here, we demonstrate how MixFunn can extract an analytical expression. Traditional neural networks produce outputs as parameterized combinations of a single nonlinear function. Attempting to write the full expression of such combinations often results in lengthy and complex formulas that are difficult to interpret or understand.

MixFunn, however, offers an advantage by restricting the range of combinations through the use of a diverse set of preselected nonlinear functions. This approach results in much shorter and more interpretable expressions. By drawing from a chosen set of nonlinear functions, MixFunn ensures that the resulting analytical expressions are more concise and human-readable. This facilitates easier understanding and analysis of the underlying relationships captured by the model.

The ability to derive more interpretable expressions is particularly valuable in scientific and engineering contexts, where understanding the functional form of the solution can provide deeper insights into the underlying phenomena. With MixFunn, researchers can gain clearer insights into the behavior of the model, making it easier to validate, interpret, and apply the results in practical scenarios.

The Mix2Funn architecture used in our experiments comprises only 77 parameters, a nearly three orders of magnitude reduction compared to the standard PINN model, which contains 66,433 parameters. Despite this highly compact design, the analytical expression derived from the MixFunn model remains relatively lengthy. To further streamline the model and reduce the complexity of the resulting expression, we apply magnitude pruning to the 35 parameters associated with the nonlinear functions. 

\begin{figure}[t]
\includegraphics[width=1.0\linewidth]{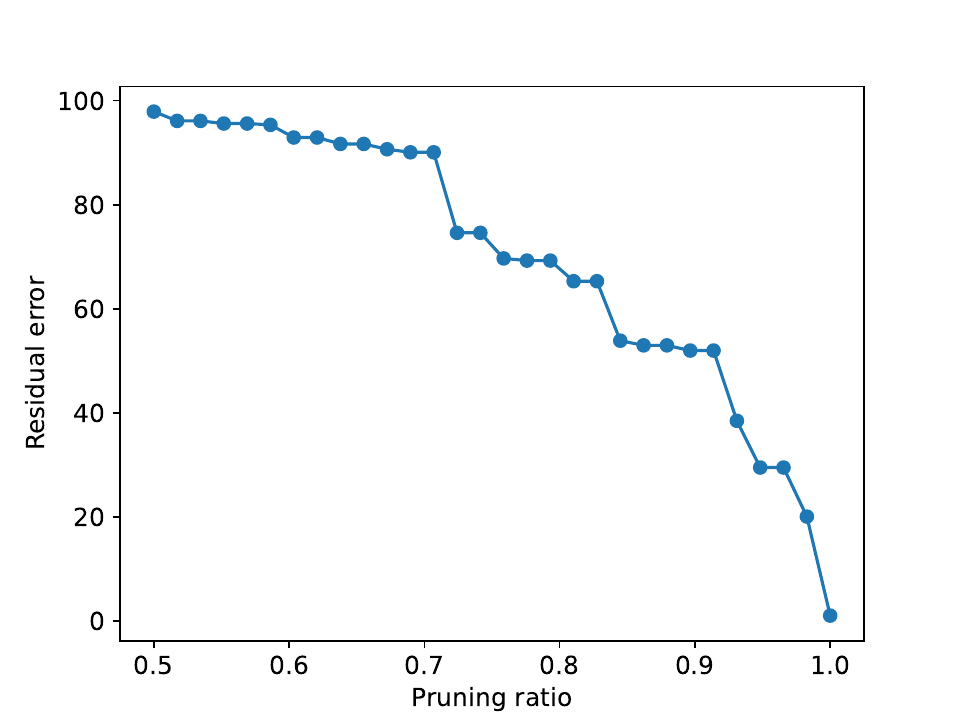}
\caption{\justifying Residual error as a function of the pruning ratio for the Mix2Funn model. The figure demonstrates that increasing the pruning ratio reduces the residual error, with the lowest error achieved at the highest pruning ratio, where all but one parameter are pruned.}
\label{fig:oscillator_pruning}
\end{figure}

The pruning ratio, a hyperparameter, determines the proportion of parameters removed during pruning. To evaluate the impact of pruning on model performance, we analyze the relationship between the pruning ratio and the residual error, which is computed as the sum of the residual loss and the initial condition loss. Figure \ref{fig:oscillator_pruning} illustrates this relationship, showing that the residual error decreases as the pruning ratio increases. The model achieves the lowest residual error at extremely high pruning ratios, where all but one parameter are pruned.

Given that the highest pruning ratio yielded the lowest residual error, we adopted this configuration to extract the analytical expression. Pruning reduced the model from 77 to 43 parameters by eliminating 34 parameters associated with the nonlinear functions. Furthermore, since many of the remaining parameters corresponded to inputs for these pruned functions, the effective parameter count was further reduced to just 5. This constitutes a reduction of four orders of magnitude, amounting to approximately $0.0075\%$ of the size of the standard PINN model. Using this pruned configuration, the following analytical expression was derived to represent the neural network's output:

\begin{equation}
	x(t) = 1.085 sin(1.006t + 1.497)e^{-|0.085t + 0.13|},
\end{equation}
which closely approximates the true analytical solution for the damped harmonic oscillator: $x_{true}(t) = sin(t + \pi/2)e^{-0.05t}$.

The derived expression demonstrates a remarkable alignment with the true solution, capturing both the oscillatory behavior and the exponential decay characteristic of the system. The similarity highlights the ability of MixFunn to effectively approximate complex differential equations while providing interpretable and concise representations. This result underscores the potential of the proposed approach not only for generating accurate solutions but also for deriving human-readable expressions that offer deeper insights into the underlying physical phenomena.

\subsubsection{Forced harmonic oscillator}

In the second harmonic oscillator case, we extend our investigation to the damped harmonic oscillator under the influence of an external driving force, $F(t) \ne 0$. Unlike the unforced case, the presence of $F(t)$ introduces a variety of dynamic behaviors, including resonance, where the oscillations are amplified if the driving frequency matches the natural frequency of the system. This configuration allows us to explore more complex phenomena such as transient and steady-state responses. Our study focuses on a specific form of $F(t)$ to examine how PINNs handle these additional complexities. 

In this study, we choose a time-dependent external force of the form $F(t)=F_0 sin(\omega t)$, where $F_0$ is the amplitude of the force and $\omega$ is its driving frequency. For frequencies far from the natural frequency of the system $\omega_0 = \sqrt{k/m}$, the system exhibits behavior similar to that of a damped harmonic oscillator, with oscillations that decay over time due to the damping effect. However, as the driving frequency $\omega$ approaches the resonance frequency $\omega_0$, the system exhibits increasing oscillation amplitudes before eventually stabilizing as the damping effect counteracts the energy input from the external force.

We focus on the behavior of the system near the resonance frequency, where more complex dynamic behaviors emerge. Specifically, we explore the case where $m=1$, $k=1$, $\gamma=0.1$, $F_0=1$ and $\omega=0.9$. It is important to note that, under these conditions, the system's resonance frequency is $\omega_0=1$.

\begin{figure*}
    \centering
    \begin{subfigure}{0.43\textwidth}
    \centering
        \includegraphics[width=1\linewidth]{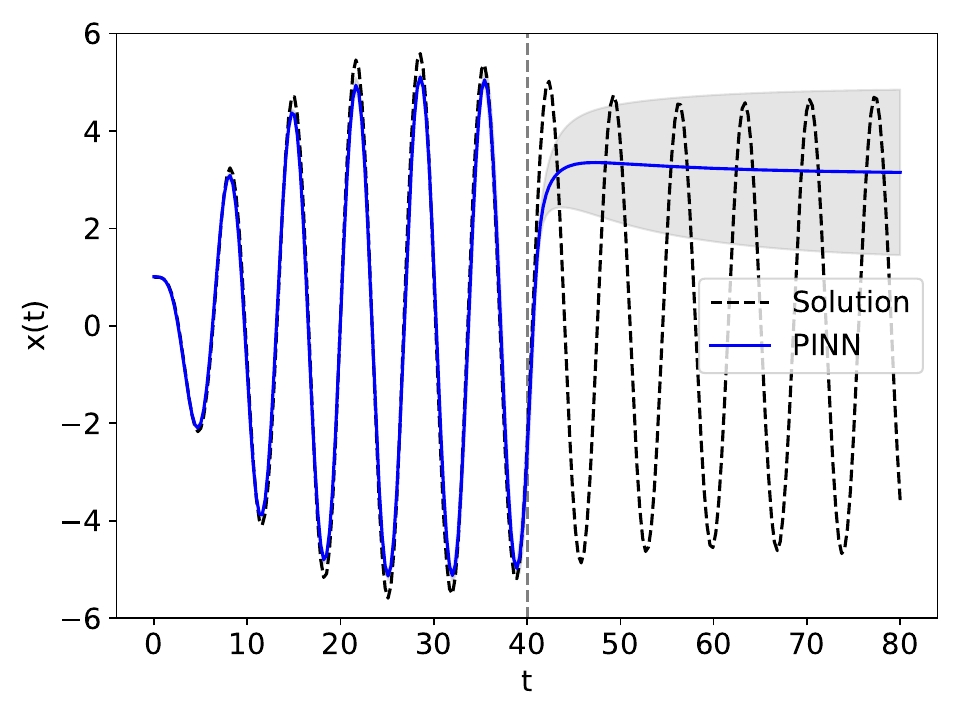}
        \caption{}
        \label{}
    \end{subfigure}%
    \begin{subfigure}{0.43\textwidth}
    \centering
        \includegraphics[width=1\linewidth]{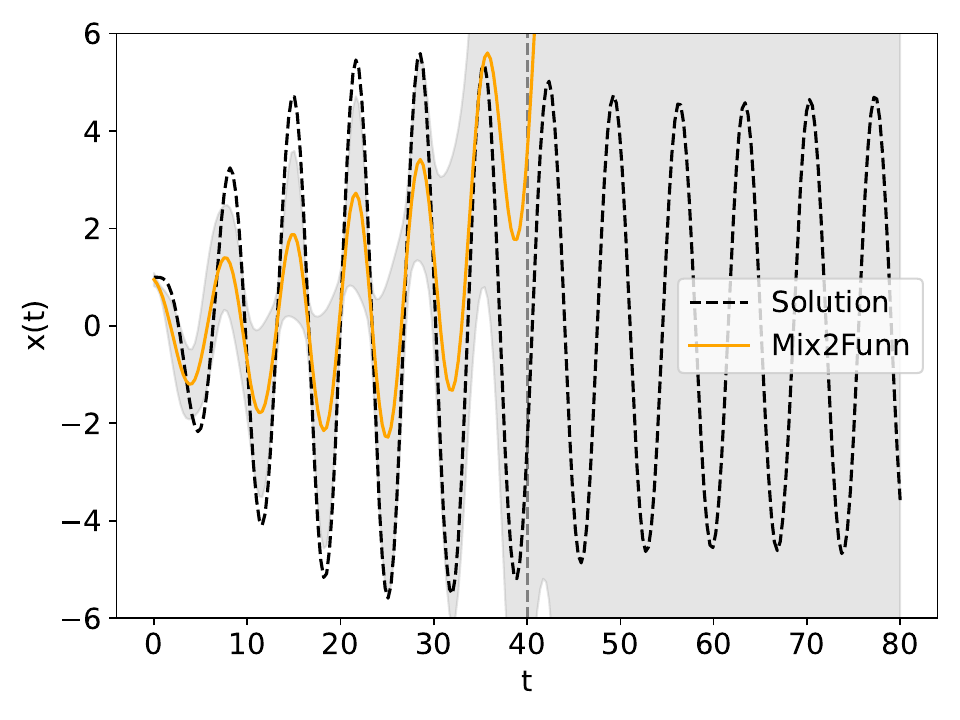}
        \caption{}
        \label{}
    \end{subfigure}
    \begin{subfigure}{0.43\textwidth}
    \centering
        \includegraphics[width=1\linewidth]{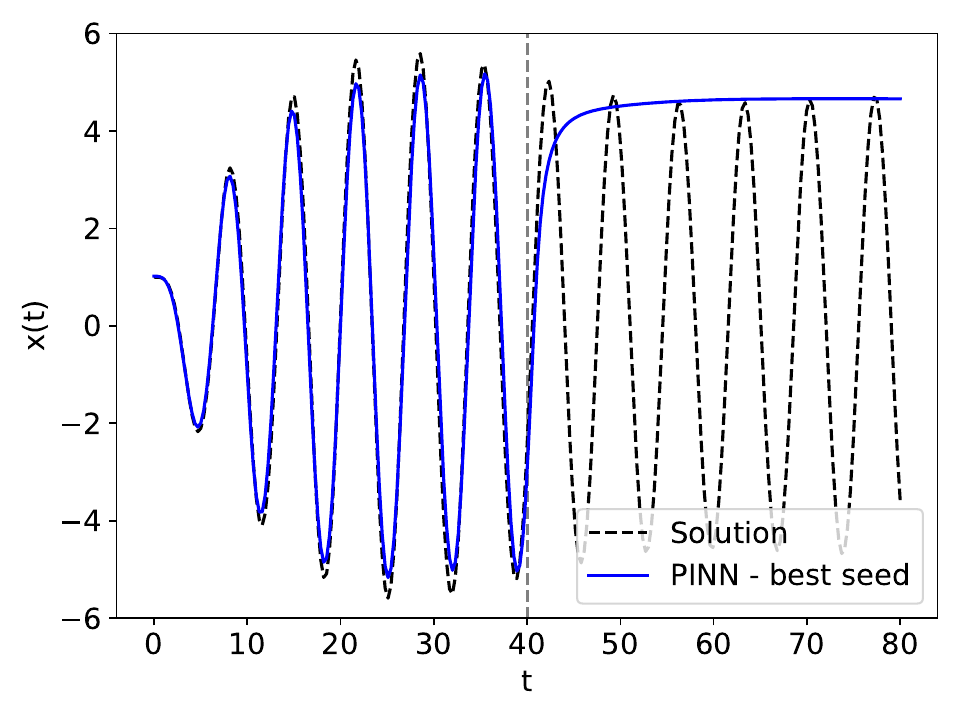}
        \caption{}
        \label{}
    \end{subfigure}%
    \begin{subfigure}{0.43\textwidth}
    \centering
        \includegraphics[width=1\linewidth]{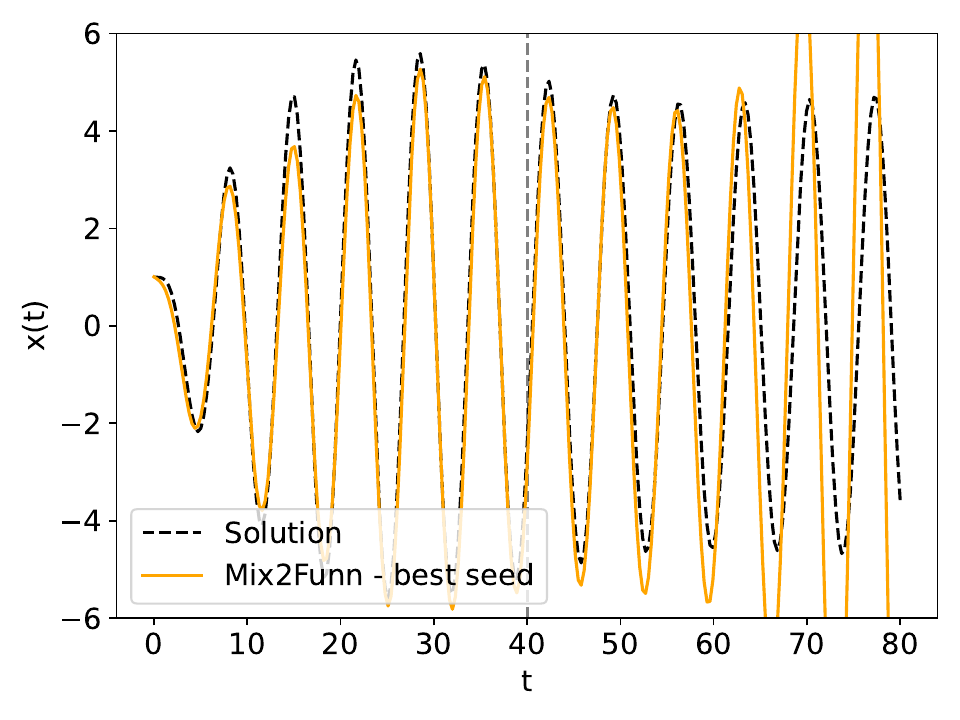}
        \caption{}
        \label{}
    \end{subfigure}
    \caption{\justifying Comparison of the predicted solutions of the PINN ((a) and (c)) and Mix2Funn ((b) and (d)) models with the exact solution, for the forced damped harmonic oscillator. In figures (a) and (b), we have the average solution in five different initializations, and the gray shade represents its variance. In figures (c) and (d), we have the best result of each model, based on the lowest total loss. While PINN achieves better accuracy in the training domain on average, Mix2Funn demonstrates comparable performance in its best initialization. In the generalization domain, PINN fails to capture the correct dynamics and shows a constant behavior for longer times, whereas Mix2Funn more accurately reflects the oscillatory nature of the system, indicating stronger generalization capabilities.}
    \label{fig:forced}
\end{figure*}

Figure \ref{fig:forced} compares the solutions predicted by the PINN and Mix2Funn models against the exact analytical solution. The results include the mean solution across five different initializations for each model and the best-performing trial, as measured by the lowest total loss. On average, the PINN achieves better solution accuracy within the training domain, while the Mix2Funn shows more variability, mainly due to the influence of poor results from some initializations. When considering the mean solution, this averaging effect results in a lower overall accuracy for Mix2Funn.

However, when focusing on the best-trained models, both PINN and Mix2Funn demonstrate similar accuracy in the training domain, with both models closely approximating the exact solution. This suggests that, given favorable initializations, the proposed Mix2Funn can achieve comparable results to PINN regarding solution quality within the training region.

The models exhibit different behaviors in the generalization domain, which lies beyond the training interval. The PINN struggles to generalize effectively, producing a near-constant output for large values of $t$. This failure to generalize is likely due to the increasing time values, which oversaturate the network and lead to poor extrapolation capabilities. In contrast, the Mix2Funn model captures the dynamics of the system more accurately, showing a better understanding of the oscillatory behavior, even for higher time values. This highlights the generalization ability of Mix2Funn, which could be particularly important for solving problems involving time-dependent systems and longer time horizons.

The comparison between the mean and best results raises the question of ensuring that the solutions found by either method are sufficiently accurate. One possible approach is to monitor the residual loss, or more precisely, the total loss, which is composed by the residual loss, initial/boundary condition loss, and, if applicable, the data loss. By tracking these loss components, we can gain insight into the performance of the model and the quality of its solution over time.

\begin{figure}[h]
    \includegraphics[width=1\linewidth]{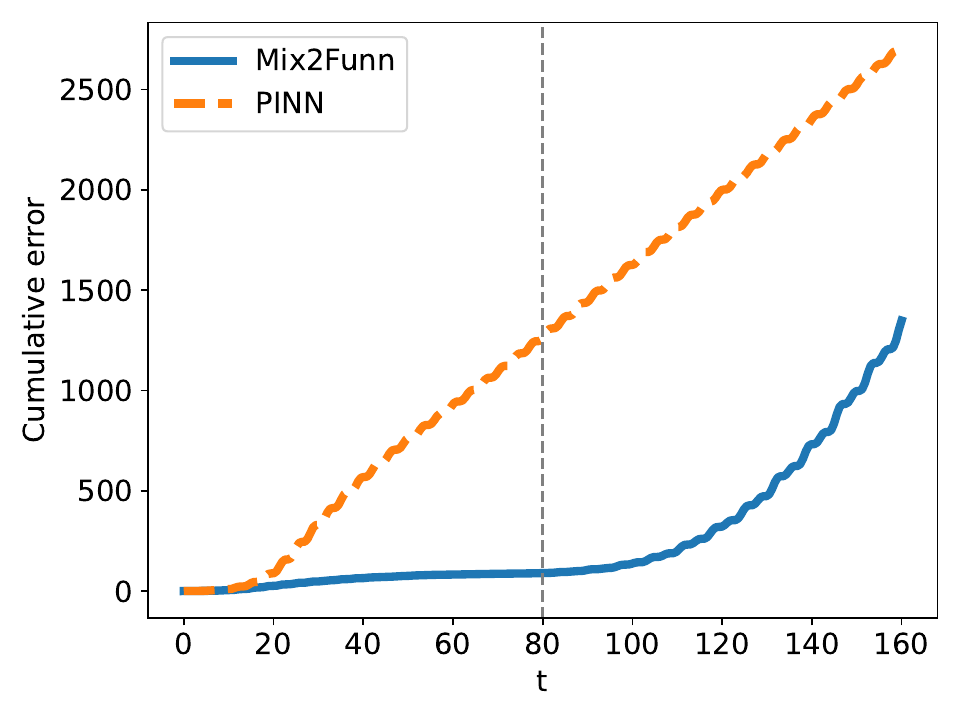}
    \caption{\justifying Cumulative error comparison between PINN and Mix2Funn models for $T_{max}=80$ on the forced damped harmonic oscillator problem. The cumulative error, representing the total accumulated error over time, is shown for both the training and generalization domains. Mix2Funn exhibits significantly slower error growth than PINN, even in the generalization domain, indicating stronger long-term accuracy and better generalization performance.}
    \label{fig:cumulative_error_forced}
\end{figure}

Given that larger training domains tend to result in better generalization, Figure \ref{fig:cumulative_error_forced} presents the cumulative error for $T_{max} = 80$. The cumulative error represents the total accumulated error across the training and generalization domains, calculated as the sum of the square differences between the predicted and true solutions over time. The results show that the cumulative error for Mix2Funn grows more slowly than for PINNs, even in the generalization domain. This indicates that Mix2Funn maintains more accurate predictions as time progresses, reflecting its stronger ability to generalize beyond the training data. In contrast, PINN exhibits a rapid increase in cumulative error, especially in the generalization domain, underscoring its limitations in handling higher time values and complex dynamics.

\subsection{Burgers equation}

The second problem to which we apply MixFunn pertains more closely to engineering applications. Specifically, we focus on solving Burgers' equation \cite{burgers1, burgers2, burgers3}, a fundamental partial differential equation in fluid dynamics. Burgers' equation serves as a prototype for various fluid dynamic phenomena due to its incorporation of both nonlinear and diffusive terms. These characteristics allow the equation to model behaviors observed in real fluid systems, such as shock waves and turbulence.
The one-dimensional form of Burgers' equation is defined as:
\begin{equation}
	\frac{\partial u}{\partial t} = -u \frac{\partial u}{\partial x} + k\frac{\partial^2 u}{\partial x^2},
\end{equation}
where $u$ represents the wave velocity and $k$ is the diffusion coefficient. The term $u \frac{\partial u}{\partial x}$ introduces nonlinearity into the equation, leading to the development of steep gradients and shock waves, while the term $k\frac{\partial^2 u}{\partial x^2}$ accounts for diffusion, smoothing out these gradients over time.

In general, Burgers' equation does not admit a closed-form analytical solution. Consequently, numerical methods are often employed to approximate its solution. To assess the accuracy of the solutions generated by the MixFunn and PINN models, we compared them against a numerical solution obtained using the fourth-order Runge-Kutta method.

Our primary objective was to identify the minimal model configuration capable of producing an approximate solution within a defined error tolerance. To achieve this, we trained and evaluated multiple model variants, varying the number of parameters in each configuration. This parameter variation enables an analysis of the trade-off between model complexity and solution accuracy, ensuring that the selected model achieves a balance between computational cost and performance.

To evaluate the ability of the model to approximate the solution to Burgers' equation and assess its generalization capabilities, we partitioned the spatiotemporal domain into distinct training and testing regions. Specifically, training data points were randomly sampled within the time interval $[0,1]$, while test data points were sampled within the extrapolated interval $[1,2]$. This separation ensures that the models are not merely interpolating within the training domain but are challenged to generalize to unseen regions of the solution space.

In comparing the performance of the PINN and MixFunn models, we observed distinct advantages for each approach. Specifically, the PINN model consistently achieved lower test domain errors, indicating better generalization capabilities for this problem. In contrast, the MixFunn model minimized training domain errors, suggesting greater precision in fitting the training data. Motivated by the potential to leverage the strengths of both models, we explored a hybrid architecture that integrates second-order neurons as the pre-activation function in the first hidden layer of multi-layer perceptron models, in a similar fashion to what was done in Ref. \cite{residual_quad}. This design enhances the MLP-based PINN models with the expressive power of second-order neurons, potentially improving training and test performance.

\begin{figure}
    \centering
    \begin{subfigure}{0.49\textwidth}
    \centering
        \includegraphics[width=1\linewidth]{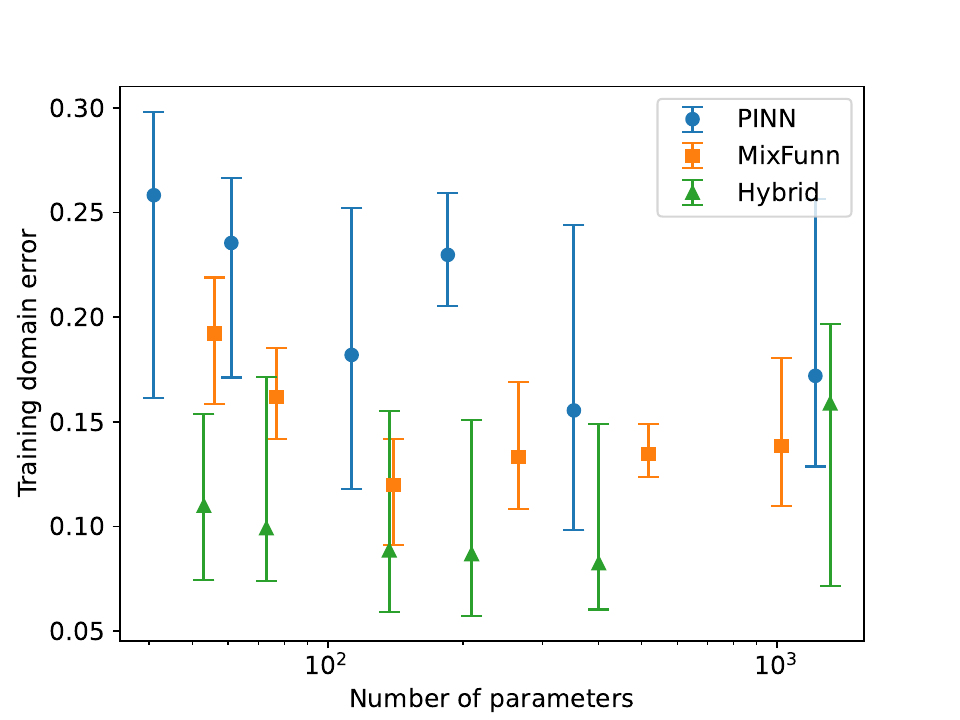}
        \caption{}
        \label{}
    \end{subfigure}%
    
    \begin{subfigure}{0.49\textwidth}
    \centering
        \includegraphics[width=1\linewidth]{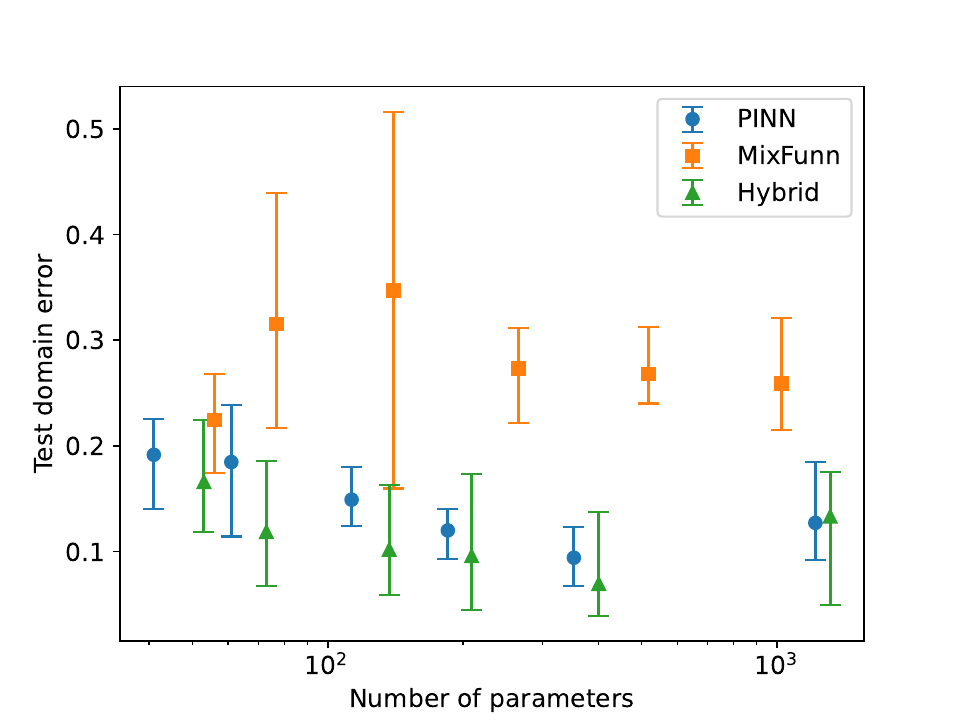}
        \caption{}
        \label{}
    \end{subfigure}
    \caption{\justifying Training (a) and test (b) domain errors for PINN, MixFunn, and hybrid models across varying parameter configurations, averaged over five independent initializations. Error bars represent the range of highest and lowest errors observed for each configuration. The results demonstrate that MixFunn achieves lower training error, PINN achieves better generalization to the test domain, and the hybrid models achieve consistently lower errors in both domains, highlighting the benefits of integrating second-order neurons into PINNs.}
    \label{fig:burgers_error}
\end{figure}

Figure \ref{fig:burgers_error} illustrates the training and test domain errors for the PINN, Mix2Funn, and hybrid models. The models were evaluated across configurations with varying numbers of parameters, and the results represent the average of five independent initializations for each configuration, along with the lowest and highest errors achieved.

The results reveal that, on average, the Mix2Funn model demonstrates higher performance in minimizing the training domain error, achieving consistently lower error values than the PINN model. Conversely, the PINN model outperforms Mix2Funn in the test domain, showcasing its better generalization to the extrapolated region. Notably, the hybrid models exhibit lower errors in both the training and test domains, indicating that incorporating second-order neurons into PINN models enhances their ability to balance training accuracy and generalization. These findings suggest that the hybrid approach effectively combines the strengths of both methodologies, providing a promising direction for further development of neural network-based solutions for differential equations.

To evaluate the performance of the models and visualize their results, it is important to note that selecting the model based solely on the lowest training domain error is not feasible. This limitation arises from the unsupervised nature of the training process, where the true solution is unknown during training and can only be assessed post hoc by comparison with alternative approaches, such as numerical methods. Consequently, the primary metric available to evaluate model quality is the residual error, which is quantified through two key components: the residual loss, representing the discrepancy in satisfying the differential equation, and the initial and boundary conditions loss, measuring the adherence of the model to these constraints.

Given these considerations, the best-performing model is selected as the one that achieves the lowest total residual error. This criterion ensures that the model approximates the underlying solution effectively and respects the physical constraints imposed by the initial and boundary conditions. By prioritizing minimizing residual error, we aim to identify models that provide accurate and physically consistent solutions to the differential equations.

\begin{figure*}
    \centering
    \begin{subfigure}{0.45\textwidth}
    \centering
        \includegraphics[width=1\linewidth]{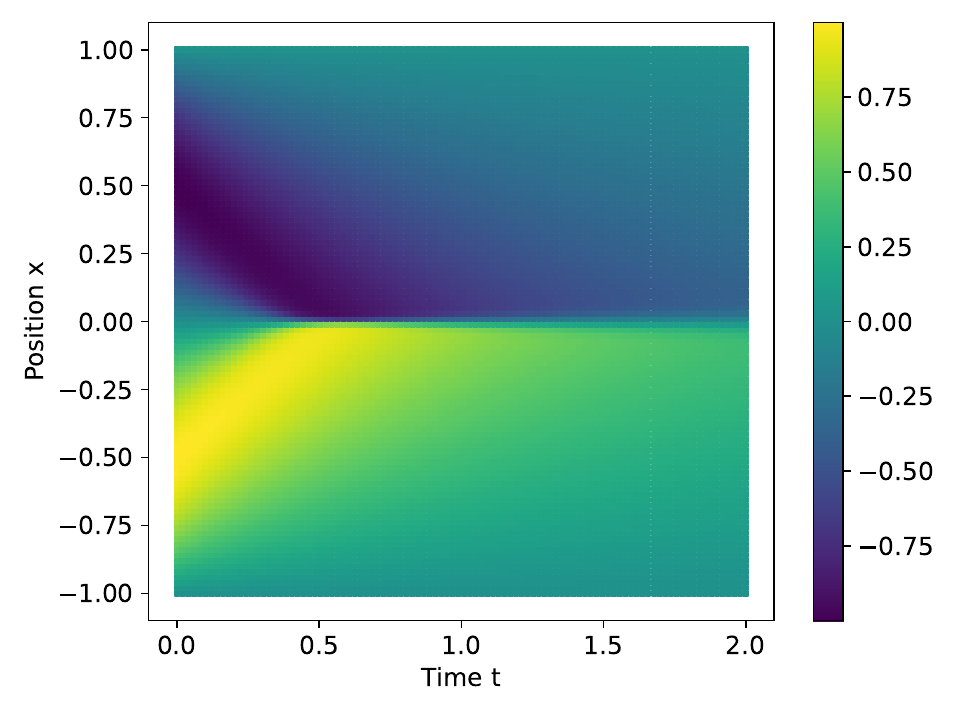}
        \caption{}
        \label{}
    \end{subfigure}%
    \begin{subfigure}{0.45\textwidth}
    \centering
        \includegraphics[width=1\linewidth]{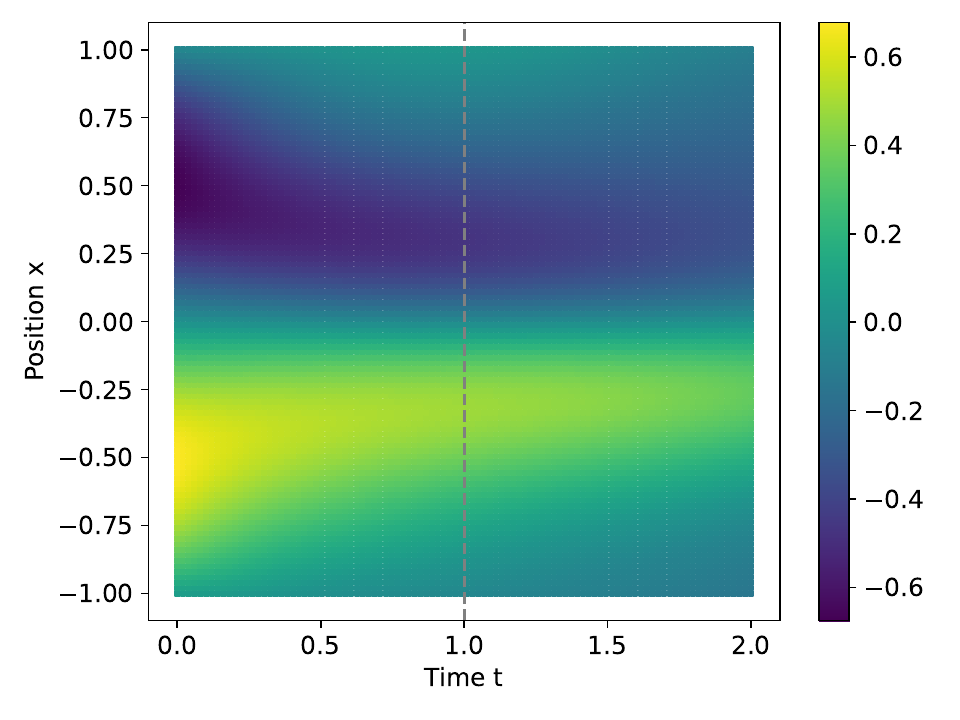}
        \caption{}
        \label{}
    \end{subfigure}
    \begin{subfigure}{0.45\textwidth}
    \centering
        \includegraphics[width=1\linewidth]{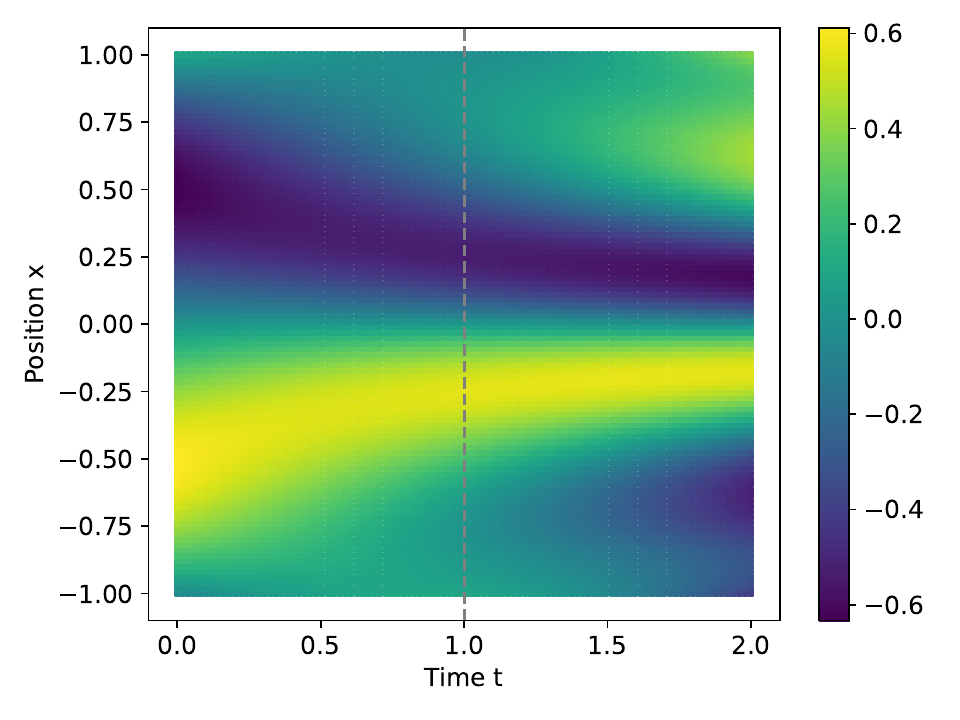}
        \caption{}
        \label{}
    \end{subfigure}%
    \begin{subfigure}{0.45\textwidth}
    \centering
        \includegraphics[width=1\linewidth]{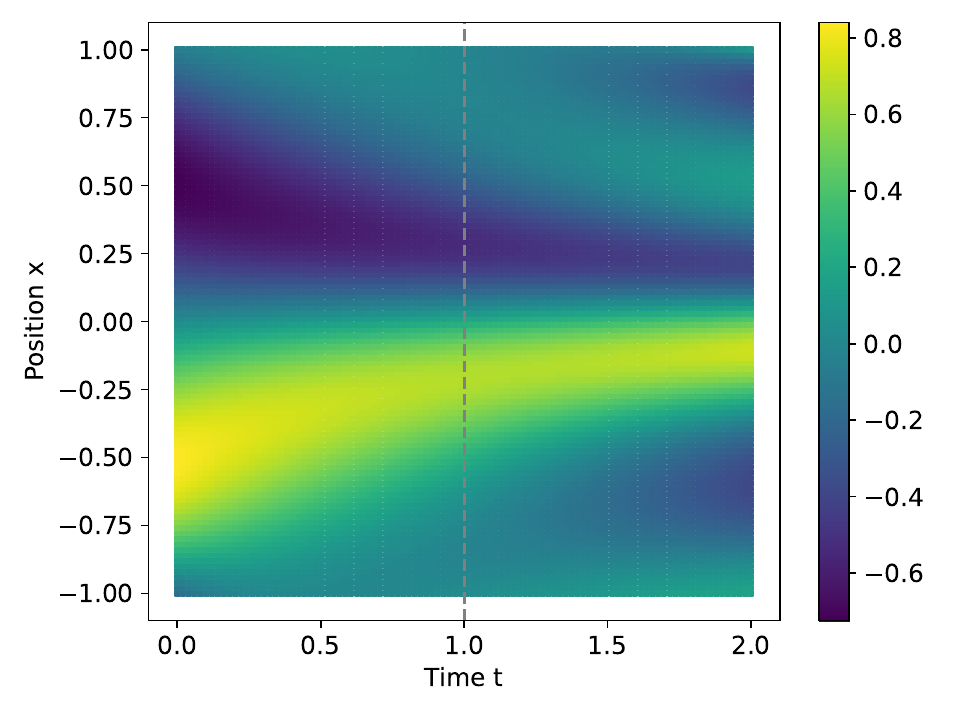}
        \caption{}
        \label{}
    \end{subfigure}
    \caption{\justifying Comparison of the (a) numerical solution for Burgers' equation with the best-performing models from the (b) PINN, (c) Mix2Funn, and (d) Hybrid architectures. The Mix2Funn and Hybrid models demonstrate a faster decay of the initial condition, aligning more closely with the numerical solution than the PINN model. The vertical dotted gray line indicates the boundary between the training and test domains.}
    \label{fig:burgers_solutions}
\end{figure*}

Figure \ref{fig:burgers_solutions} presents the best-performing model from each architecture alongside the numerical solution for Burgers' equation. While the solutions exhibit qualitatively similar behaviors, notable differences can be observed. The Mix2Funn and Hybrid models demonstrate a faster decay of the initial condition compared to the PINN model, aligning more closely with the numerical solution. 

Extracting the analytical expression from the Mix2Funn network required a more intricate process. First, we selected the trained model that achieved the lowest residual error. Next, we systematically searched for the optimal pruning ratio, following the same approach used for the damped harmonic oscillator. In this case, the model with the lowest residual error had $0\%$ pruning, meaning that all 56 parameters were retained. However, to further simplify the model, we explored fine-tuning with nonzero pruning ratios.

We identified the second and third lowest residual errors at $20\%$ and $80\%$ pruning, respectively. To balance complexity and accuracy, we selected the $80\%$ pruned model and fine-tuned it for an additional $1,000$ epochs, saving the parameters corresponding to the lowest training error during this phase. Subsequently, we continued fine-tuning with a $90\%$ pruning ratio, following the same selection procedure. This iterative process ultimately yielded a highly simplified model with only four parameters. From this optimized network, we extracted the following analytical expression:
\begin{equation}
	u(x,t) = 0.971 cos(-1.123sin(\pi x) t + 3.160x  + 1.572).
\end{equation}

\begin{table}
\begin{center}
\begin{tabular}{ | m{5em} | m{5em} | m{2cm}| m{1cm} | m{1.0cm} |} 
  \hline
   Model & Parameters & Residual error & Train error & Test error\\ 
  \hline
  \hline
  PINN & 353 & \textbf{0.123} & 0.202 & \textbf{0.089} \\
  \hline
  Mix2Funn & 56 & 0.155 & 0.218 & 0.215 \\
  \hline
  Hybrid & 73 & 0.182 & 0.171 & 0.185 \\
  \hline
  Pruned & \textbf{4} & 0.727 & \textbf{0.144} & 0.410 \\
  \hline
\end{tabular}
\end{center}
\caption{\justifying Quantitative comparison of the best-performing models from the PINN, Mix2Funn, Hybrid, and pruned architectures. Metrics include the test domain error, the training domain error, and the number of parameters for each model. The results highlight the pruned Mix2Funn with the lowest parameter count and strong performance, and the Hybrid model's balance of improved generalization and reduced parameter complexity compared to the PINN.}
\label{table:burgers}
\end{table}

Table \ref{table:burgers} presents a quantitative comparison of the models. While the PINN model achieves the lowest test error, the visualization in Figure \ref{fig:burgers_solutions} suggests that this outcome is influenced by the relatively small velocity values in the test domain rather than a superior representation of the overall solution behavior. In contrast, the Mix2Funn model closely aligns with the numerical solution while requiring fewer parameters, demonstrating its efficiency and expressive power.

Additionally, the pruned Mix2Funn model, despite exhibiting a higher residual error compared to other models, achieved the lowest training error, indicating the closest approximation to the numerical solution within the training domain, highlighting the trade-off between complexity and generalization. Meanwhile, the hybrid model exhibited an intermediate performance, balancing accuracy and parameter efficiency by compromising between the strengths of the other models.

\subsection{Quantum infinite square well}

\begin{figure*}
    \centering
    \begin{subfigure}{1.0\textwidth}
    \centering
        \includegraphics[width=1\linewidth]{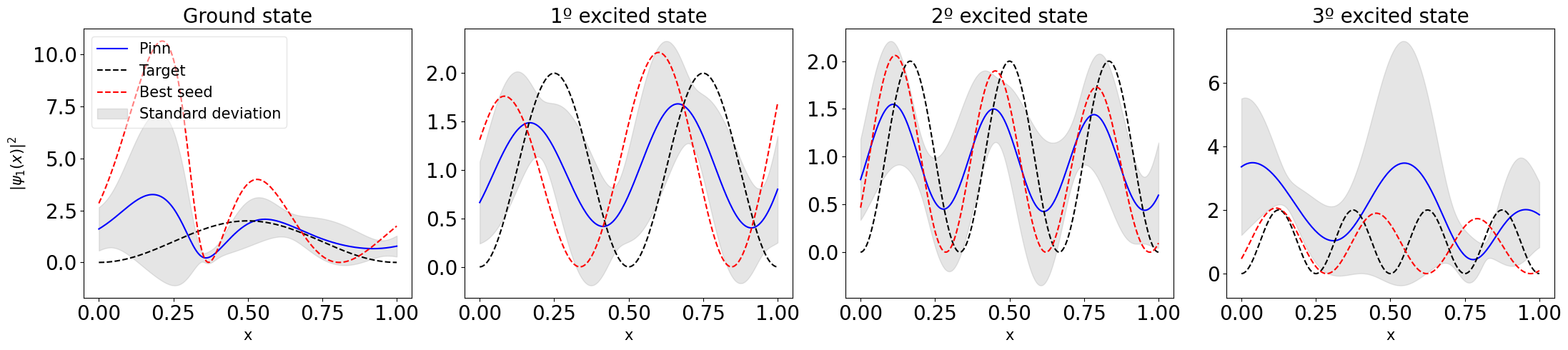}
        \caption{}
        \label{}
    \end{subfigure}%
    
    \begin{subfigure}{1.0\textwidth}
    \centering
        \includegraphics[width=1\linewidth]{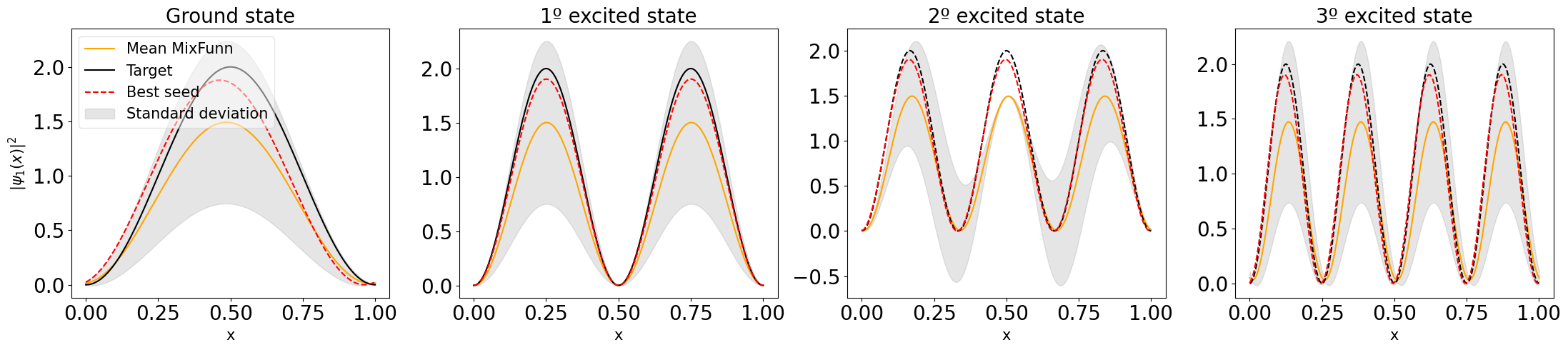}
        \caption{}
        \label{}
    \end{subfigure}
    \caption{\justifying Predicted wavefunctions for the infinite quantum well problem are shown for both the (a) PINN and (b) Mix2Funn models. Results are presented for four energy levels: the first and second excited states, which were used during training, and the ground and third excited states, which test generalization. While the PINN model provides reasonable approximations for the training states, it exhibits shifted and less accurate predictions for unseen energy levels. In contrast, the Mix2Funn model delivers more precise predictions for the training states and demonstrates superior generalization to energy levels not encountered during training.}
    \label{fig:solution_well_pinn}
\end{figure*}

The time-independent Schrödinger equation relates to solving an eigenvalue problem, represented by the expression $Av=av$ for some matrix $A$, eigenvector $v$, and eigenvalue $a$. Specifically, in the context of quantum mechanics, the equation is written as:
\begin{equation}
	H \psi = E\psi,
\end{equation}
where $E$ represents the energy eigenvalue, $\psi$ is the quantum eigenfunction corresponding to the energy $E$, and $H$ is the Hamiltonian operator, which is typically known from the physical problem under consideration. The non-relativistic time-independent Schrödinger equation for a particle with mass can be formulated as the following differential equation:
\begin{equation}
	-\frac{\hbar^2}{2m}\frac{d^2}{dx^2}\psi + V(x)\psi = E\psi,
\end{equation}
where $\hbar$ denotes the reduced Planck constant, $m$ is the mass of the particle, and $V(x)$ represents the potential energy as a function of position $x$.

One significant challenge with this equation is that it necessitates prior knowledge of the energy eigenvalue $E$ to find a solution. It is often possible to determine the energy eigenvalues analytically for simple potential functions. However, for more complex potentials, analytical solutions may not be feasible. In such cases, numerical methods and eigenvalue techniques become essential to solve the equation.

Here we propose an alternative approach to finding the eigenvalues of the Schrödinger equation by leveraging neural networks, specifically PINN or MixFunn. The traditional eigenvalue problem $Av=av$ holds true only for specific eigenvalues $a$. By utilizing neural networks, we can effectively search for these eigenvalues through a process of training and loss evaluation.

In this approach, we generate a range of candidate eigenvalues within an interval. We train a neural network for each candidate eigenvalue to approximate the corresponding wavefunction. The neural network is trained for a fixed number of epochs for each candidate eigenvalue, and the loss function is calculated based on the residuals of the differential equation. We compare the resulting loss values after training the neural network for each candidate eigenvalue. The correct eigenvalues are identified as those that correspond to the lowest loss values, indicating that the neural network has successfully approximated the solution to the differential equation.

By using this method, we avoid the need to update the eigenvalue estimates within the training process iteratively. Instead, we evaluate a discrete set of potential eigenvalues, training the neural network separately for each one, and then select the eigenvalues that yield the minimum loss. This approach transforms the problem of finding eigenvalues into a series of training and evaluation steps, leveraging the capability of the neural network to approximate the solution of a differential equation.

To illustrate this, we apply the Mix2Funn framework to solve the infinite quantum well problem \cite{griffiths_introduction_2018}.
The potential $V(x)$ for the infinite quantum well is defined as:
\begin{equation}
	V(x) = 
	\begin{cases}
		\infty, \mbox{ if } |x| > 1, \\
		0, \mbox{ otherwise}.
	\end{cases}
\end{equation}

To avoid the computational challenges associated with handling infinite potential values in numerical simulations, we impose boundary conditions such that $\psi(-1)=\psi(1)=0$. This choice is based on the physical requirement that the wave function must vanish at the boundaries where the potential is infinite.

\begin{figure}
    \includegraphics[width=1\linewidth]{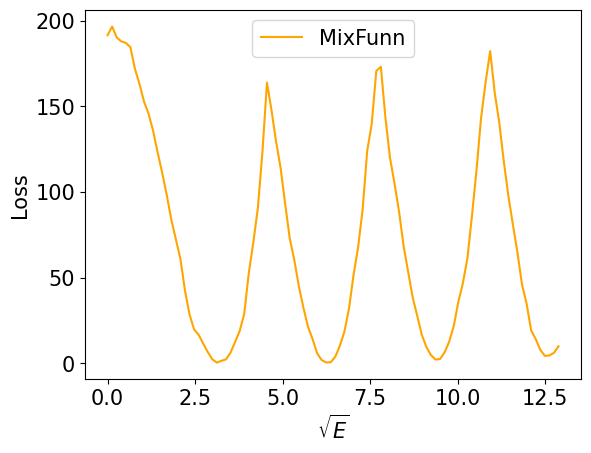}
    \caption{\justifying Loss as a function of candidate square-root energy eigenvalues for the infinite quantum well problem. Each point represents the lowest training loss of a neural network model evaluated at a fixed candidate $\sqrt{E}$. The minima of the curve correspond to the most approximated $\sqrt{E}$ of the system, where the neural network best satisfies the Schrödinger equation.}
    \label{fig:loss_energy}
\end{figure}

We use the optimization search method described previously to determine the system's correct, or most approximate, eigenvalues, as shown in Figure \ref{fig:loss_energy}. By generating a range of candidate eigenvalues and training the neural network for each value, we evaluate the loss function, which measures the residuals of the Schrödinger equation. The eigenvalues corresponding to the lowest loss values are identified as the most accurate solutions.

In the context of this problem, generalization of the model refers to its ability to accurately predict wavefunctions corresponding to different energy values beyond those used during training. To evaluate this capability, the model is designed to accept two inputs: the position $x$ and the square root of the energy $\sqrt{E}$\footnote{Although alternative representations such as using $E$ directly were considered, empirical experiments revealed that using $\sqrt{E}$ yielded better performance.}. To promote generalization, the models were trained using eigen-energies corresponding to the first two excited states, sampled randomly during training. For the MixFunn model, the training procedure involved an initial phase of 10,000 epochs, followed by a pruning step as outlined in previous sections. Subsequently, the pruned model was fine-tuned for 10,000 epochs.

Figure \ref{fig:solution_well_pinn} shows the predicted wavefunctions obtained using the Mix2Funn and PINN models. The results are presented for four distinct energy eigenvalues: two corresponding to the states used during training (the first and second excited states) and two additional states (the ground and third excited states) employed to assess the generalization capabilities of the models. The Mix2Funn model, evaluated using both mean and best-seed performance, demonstrates better learning of the wavefunctions and exhibits improved generalization to unseen energy levels. In contrast, the PINN model yields shifted approximations for the trained energies and shows limited accuracy when predicting wavefunctions for energies outside the training set.

\begin{figure}
    \centering
    \includegraphics[width=1\linewidth]{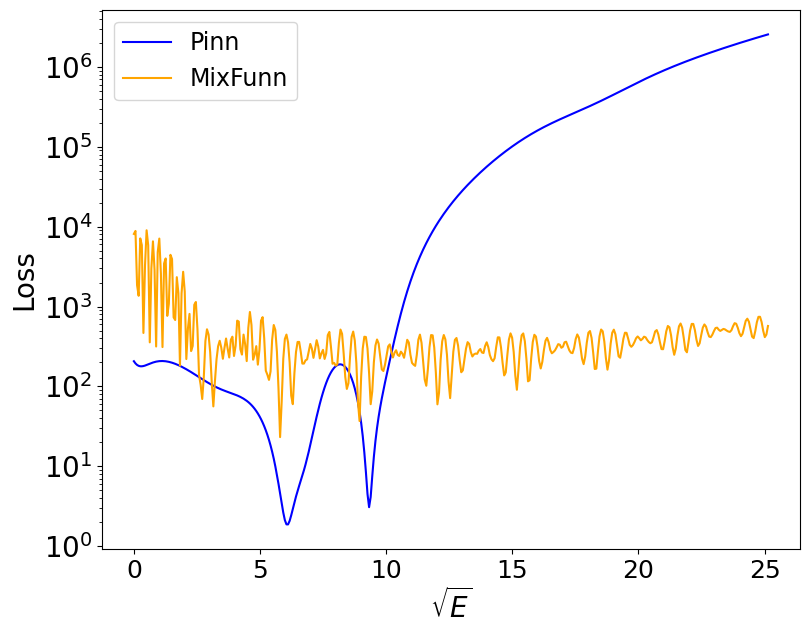}
    \caption{\justifying Lowest training error as a function of the square root of the energy level for the best trained PINN and Mix2Funn models trained only with the second and third energy levels. The PINN model achieves low error exclusively at the energy levels used during training, indicating limited generalization. Conversely, the Mix2Funn model identifies multiple loss minima that align closely with the correct $\sqrt{E}$, even for energy levels outside the training set, highlighting its enhanced generalization capability.}
    \label{fig:loss_energy_singlemodel}
\end{figure}

Another approach to assess the generalization capability of the model is by plotting the lowest training error as a function of different energy values. This approach is analogous to the method used for identifying eigenvalues, but is applied here to a single model trained on the first two excited states. Figure \ref{fig:loss_energy_singlemodel} presents this error for the PINN and Mix2Funn models. The PINN model achieves low error values only for the energy levels used during training, indicating limited generalization. In contrast, the Mix2Funn model identifies distinct loss minima that align closely with the correct eigenvalues, including those outside the training set, demonstrating better generalization capability.

To derive the analytical expression predicted by the model, we extract it following the pruning and fine-tuning phases. This approach removes redundant or unnecessary parameters, resulting in a more compact and interpretable expression. By focusing on the pruned and fine-tuned model, we balance model complexity and accuracy, preserving only the essential parameters necessary for representing the solution. For this analysis, we selected the parameter values corresponding to the model that achieved the lowest training error among the five independent training seeds. The analytical expression obtained is: 
\begin{equation}
	\psi(x,E) = 1.38\sin(0.9998\sqrt{E}\cdot x + 8.01\sqrt{E} +9.359),
\end{equation}
which closely approximates the true analytical solution $ \psi_{true}(x,E) = \sqrt{2}\sin(\sqrt{E}x) $, with a phase shift given by $ 8.01\sqrt{E} +9.359 $.

\section{Conclusions}

We introduced MixFunn, a physics-informed neural network architecture specifically designed to approximate solutions to differential equations. The architecture integrates two key components: mixed-function neurons, which combine various parameterized nonlinear functions to enhance representational flexibility, and second-order neurons, which incorporate a linear transformation of their inputs alongside quadratic terms that capture input cross-combinations.

The design of this architecture is inspired by the common forms of solutions to differential equations encountered in physics, where trigonometric and exponential functions, often coupled with cross-combined terms, frequently appear. By embedding this inductive bias through carefully selecting nonlinear functions, the model structure is aligned with the expected solution space, promoting efficiency and accuracy.

We evaluated MixFunn on four distinct problems: the damped and forced harmonic oscillator, Burgers' equation, and the quantum infinite well. We used a fully unsupervised training regime based solely on residual, boundary, and initial condition losses. Despite the inherent challenges of training without explicit data, MixFunn demonstrated robust performance across all test cases.

To further enhance generalization, we incorporated a suite of regularization techniques, including dropout, pruning, and function normalization. These methods encourage competition among different nonlinear functions and neurons, effectively reducing loss while simplifying the resulting models. Notably, our approach drastically reduced the number of parameters, up to four orders of magnitude fewer than conventional PINN architectures, without compromising performance. This significant reduction in model complexity decreases computational overhead and facilitates the extraction of interpretable analytical expressions from the trained models.

Our experimental results indicate that MixFunn consistently outperforms traditional PINN architectures based on multi-layer perceptrons, delivering superior accuracy and generalization in extrapolated domains. Moreover, hybrid models that enhance standard PINN architectures with second-order neurons exhibited improved performance with fewer parameters, further underscoring the potential of second-order neurons to boost expressivity while maintaining efficiency.

We believe that MixFunn represents a significant advancement in physics-informed neural networks, offering a robust framework for solving complex differential equations. Its ability to combine high accuracy, interpretability, and parameter efficiency makes it a versatile tool for various scientific and engineering applications. Future work could explore the extension of MixFunn to other domains, such as stochastic differential equations or systems with high-dimensional input spaces, further showing its utility in solving challenging computational problems in physics and beyond.

\section*{Data availability}

The code for the MixFunn architecture, along with the examples shown in this article, can be found at \url{https://github.com/tiago939/MixFunn}.

\vspace{10pt}

\section*{Acknowledgments}

This work was supported by the S\~ao Paulo Research Foundation (FAPESP), under
Grant No. 2023/15739-3 and No. 2022/00209-6, by the National Council for Scientific and Technological Development (CNPq), under Grants No. 140468/2022-6,  No. 309862/2021-3, No. 409673/2022-6, No. 311612/2021-0, and No. 421792/2022-1, and by the National Institute for the Science and Technology of Quantum Information (INCT-IQ) under Grant No. 465469/2014-0.
\bibliography{references}

\end{document}